\documentclass[twoside]{article}

\usepackage[accepted]{aistats2023}
\usepackage[numbers, compress]{natbib}
\usepackage[dvipsnames]{xcolor}
\usepackage{colortbl}

\usepackage{bm}
\usepackage{amsfonts}
\usepackage{amsmath}
\usepackage{amsthm}

\usepackage{booktabs} 
\usepackage{wrapfig}

\newtheorem{lemma}{Lemma}
\newtheorem{proposition}{Proposition}

\newtheorem{corollary}{Corollary}

\usepackage[utf8]{inputenc} 
\usepackage[T1]{fontenc}    
\usepackage{hyperref}       
\usepackage{url}            
\usepackage{booktabs}       

\usepackage{nicefrac}       
\usepackage{microtype}      
\usepackage{graphicx}       
\usepackage{subcaption}     

\def\to{{\,\rightarrow\,}}

\mathchardef\mhyphen="2D

\newcommand{\vertiii}[1]{{\left\vert\kern-0.25ex\left\vert\kern-0.25ex\left\vert #1
    \right\vert\kern-0.25ex\right\vert\kern-0.25ex\right\vert}}

\def\bd{{\mathbf{d}}}

\def\bv{{\mathbf{v}}}
\def\bw{{\mathbf{w}}}
\def\bx{{\mathbf{x}}}

\def\bz{{\mathbf{z}}}
\def\bA{{\mathbf{A}}}
\def\bB{{\mathbf{B}}}
\def\bC{{\mathbf{C}}}
\def\bD{{\mathbf{D}}}

\def\bH{{\mathbf{H}}}
\def\bI{{\mathbf{I}}}
\def\bJ{{\mathbf{J}}}

\def\bM{{\mathbf{M}}}

\def\bY{{\mathbf{Y}}}

\usepackage{graphicx}
\usepackage{epstopdf}
\usepackage{subcaption}
\usepackage{multirow}

\usepackage{tabu}
\usepackage{tikz}
\usepackage{pgfplots}
\usetikzlibrary{shapes.geometric,arrows,arrows.meta, fit}

\usepackage{algorithm}
\usepackage{algorithmic}

\newcommand{\xC}{\check{\mathbf{x}}}
\newcommand{\zC}{\check{\mathbf{z}}}
\newcommand{\bdelta}{\bm{\delta}}

\newcommand{\upmodels}{\perp\!\!\!\perp} 

\DeclareMathOperator*{\argmin}{arg\,min}

\usepackage{tikz}
\usepackage{pgfplots}
\usetikzlibrary{shapes.geometric,arrows,arrows.meta, fit}

\hypersetup{
colorlinks = true,
linkcolor = Blue,
anchorcolor = blue,
citecolor = Blue,
filecolor = cyan,
menucolor = ForestGreen,
runcolor = cyan,
urlcolor = ForestGreen}

\title{Decomposing Counterfactual Explanations for Consequential Decision Making}

\begin{document}

%
\runningtitle{Decomposing Counterfactual Explanations}

\twocolumn[

\aistatstitle{Decomposing Counterfactual Explanations \\ for Consequential Decision Making}

\aistatsauthor{Martin Pawelczyk \And Lea Tiyavorabun \And  Gjergji Kasneci}

\aistatsaddress{University of Tübingen \And  University of Amsterdam \And University of Tübingen} ]

\begin{abstract}
The goal of algorithmic recourse is to reverse unfavorable decisions (e.g., from loan denial to approval) under automated decision making by suggesting actionable feature changes (e.g., reduce the number of credit cards).
To generate low-cost recourse the majority of methods work under the assumption that the features are independently manipulable (IMF).
To address the feature dependency issue the recourse problem is usually studied through the causal recourse paradigm.
However, it is well known that strong assumptions, as encoded in causal models and structural equations, hinder the applicability of these methods in complex domains where causal dependency structures are ambiguous.
In this work, we develop \texttt{DEAR} (DisEntangling Algorithmic Recourse), a novel and practical recourse framework that bridges the gap between the IMF and the strong causal assumptions.
\texttt{DEAR} generates recourses by disentangling the latent representation of co-varying features from a subset of promising recourse features to capture the main practical recourse desiderata.
Our experiments on real-world data corroborate our theoretically motivated recourse model and highlight our framework's ability to provide reliable, low-cost recourse in the presence of feature dependencies. 

\end{abstract}

\section{Introduction} \label{section:introduction}
Counterfactual explanations provide a means for actionable model explanations at feature level. 
Such explanations, which have become popular among legal and technical communities, provide both an explanation and an instruction: 
the former emphasizes why a certain machine learning (ML) prediction was produced;
the latter gives an instruction on how to act to arrive at a desirable outcome. 


Several approaches in recent literature tackled the problem of providing recourses by generating counterfactual explanations \citep{wachter2017counterfactual,ustun2019actionable}. 
For instance, \citet{wachter2017counterfactual} proposed a gradient based approach which finds the nearest counterfactual resulting in the desired prediction. 
\citet{pawelczyk2019} proposed a method which uses a generative model to find recourses in dense regions of the input space.
More recently, \citet{karimi2020intervention} advocated for considering causal structure of the underlying data when generating recourses to avoid spurious explanations. 
Yet, despite their popularity these works are not without drawbacks: (i) \citet{wachter2017counterfactual} implicitly assume that the input features can be independently manipulated, (ii) \citet{pawelczyk2019} narrowly focuse on generating recourse in dense regions of the input space, and (iii) \citet{karimi2020intervention} require a correct specification of the causal graph and structural equation models.

For many practical use cases the strong causal assumptions constitute the limiting factor when it comes to the deployment of causal recourse methods.
On the other hand, most of the practical approaches implicitly make the \emph{independently manipulable feature} (IMF) assumption ignoring feature dependencies. 
Therefore critiques of counterfactual explanations and algorithmic recourse have highlighted the feature dependency issue \citep{barocas2020,Venkatasubramanian2020}: in a nutshell, \emph{changing one feature will likely change others}.
For instance, a recourse system might ask to increase the feature `income' for a loan approval. 
However, there might be several ways of achieving the same desired outcome of loan approval: either one could increase `income' through a promotion or one could find a new role in a different company. 
In the former case, the value of the variable reflecting `time on job' would go up, which would likely amplify the model's output towards the desirable outcome. 
In the latter case, however, the model's output would likely swing towards a loan rejection, since the short `time on job' opposes the positive influence of the `income' increase. 

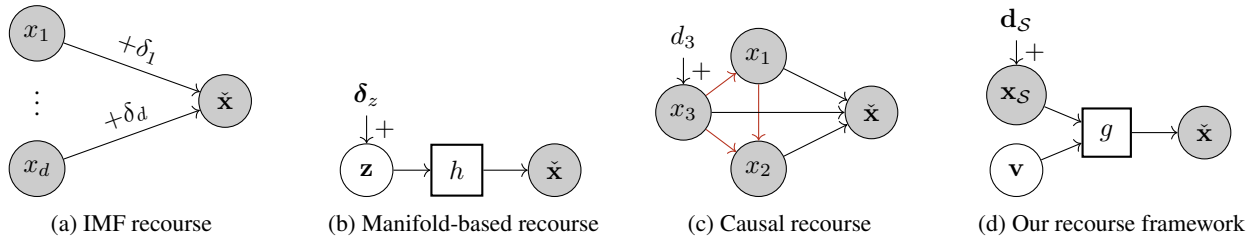
\begin{figure*}[tb]
\centering
\begin{subfigure}[b]{0.24\textwidth}
\centering

\begin{tikzpicture}[square/.style={regular polygon,regular polygon sides=4}]

\usetikzlibrary{fit}

\tikzstyle{circle1} = [circle, minimum size = 4mm, text centered, draw=white, fill=red!0]

\tikzstyle{circle2} = [circle, minimum size = 7mm, text centered, draw=black, fill=black!20]

\node(x1)[circle2, xshift=-1cm, yshift=+0.90cm] {$x_{1}$};
\node(xj)[circle1, xshift=-1cm, yshift=+0.08cm] {$\vdots$};
\node(xd)[circle2, xshift=-1cm, yshift=-0.90cm] {$x_d$};

\node(c)[circle2, xshift=+1.5cm, yshift=+0.0cm] {$\xC$};

\draw[->]  (x1) -- (c) node [midway, above, sloped] (perturb) {$+ \delta_{1} $};
\draw[->]  (xd) -- (c) node [midway, above, sloped] (perturb) {$+ \delta_{d} $};

\end{tikzpicture}
\caption{IMF recourse}
\label{fig:imf_intervention}
\end{subfigure}
\hfill
\begin{subfigure}[b]{0.24\textwidth}
\centering

\begin{tikzpicture}[square/.style={regular polygon,regular polygon sides=4}]

\usetikzlibrary{fit}

\tikzstyle{circle1} = [circle, minimum size = 7mm, text centered, draw=black, fill=red!0]

\tikzstyle{circle2} = [circle, minimum size = 7mm, text centered, draw=black, fill=black!20]

\node(delta)[xshift=-1cm, yshift=+1.0cm] {$\bdelta_z$};
\node(z)[circle1, xshift=-1cm, yshift=+0.0cm] {$\bz$};
\node (g) [square, draw, thick, xshift=+0.20cm, yshift=+0.0cm] {$h$};
\node(c)[circle2, xshift=+1.5cm, yshift=+0.0cm] {$\xC$};

\draw[->] (delta) -- (z) node [midway, above, sloped] (perturb) {$+$};
\draw[->]  (z) -- (g);
\draw[->]  (g) -- (c);

\end{tikzpicture}
\caption{Manifold-based recourse}
\label{fig:manifold_intervention}
\end{subfigure}
\begin{subfigure}[b]{0.24\textwidth}
\centering

\begin{tikzpicture}[square/.style={regular polygon,regular polygon sides=4}]

\usetikzlibrary{fit,arrows,decorations.markings}

\tikzstyle{circle1} = [circle, minimum size = 7mm, text centered, draw=black, fill=black!20]

\tikzset{
  double arrow/.style args={#1 colored by #2 and #3}{
    -stealth,line width=#1,#2, 
    postaction={draw,-stealth,#3,line width=(#1)/3,
                shorten <=(#1)/3,shorten >=2*(#1)/3}, 
  }
}


\node(delta)[xshift=-1cm, yshift=+0.25cm] {$d_{3}$};
\node(a)[circle1] {$x_1$};
\node(b)[circle1, xshift=+0.0cm, yshift=-1.5cm] {$x_2$};
\node(c)[circle1, xshift=-1.0cm, yshift=-0.75cm] {${x_3}$};
\node(xc)[circle1, xshift=+1.5cm, yshift=-0.75cm] {$\xC$};


\draw[->, BrickRed]  (a) -- (b);
\draw[->, BrickRed]  (c) -- (b);
\draw[->, BrickRed]  (c) -- (a);

\draw[->]  (a) -- (xc);
\draw[->]  (b) -- (xc);
\draw[->]  (c) -- (xc);
\draw[->] (delta) -- (c) node [midway, above, sloped] (perturb) {$+$};




\end{tikzpicture}
\caption{Causal recourse}
\label{fig:causal_intervention}
\end{subfigure}
\hfill
\begin{subfigure}[b]{0.24\textwidth}
 \centering

\begin{tikzpicture}[square/.style={regular polygon,regular polygon sides=4}]

\usetikzlibrary{fit}

\tikzstyle{circle1} = [circle, minimum size = 7mm, text centered, draw=black, fill=red!0]

\tikzstyle{circle2} = [circle, minimum size = 7mm, text centered, draw=black, fill=black!20]

\node(delta)[xshift=-1cm, yshift=+1.5cm] {$\bd_{{\mathcal{S}}}$};
\node(x)[circle2, xshift=-1cm, yshift=+0.5cm] {$\bx_{\mathcal{S}}$};
\node(v)[circle1, xshift=-1cm, yshift=-0.5cm] {$\bv$};
\node (g) [square, draw, thick, xshift=+0.20cm] {$g$};
\node(c)[circle2, xshift=+1.5cm] {$\xC$};

\draw[->] (delta) -- (x) node [midway, above, sloped] (perturb) {$+$};
\draw[->]  (x) -- (g);
\draw[->]  (v) -- (g);
\draw[->]  (g) -- (c);


\end{tikzpicture}
 \caption{Our recourse framework}
 \label{fig:our_intervention}
\end{subfigure}
\caption{\textbf{The spectrum of recourse frameworks}. Illustrating the different assumptions underlying each recourse framework. (a): Recourses are found by neglecting input dependencies (e.g., \citep{wachter2017counterfactual}).
(b): Actions made to the latent code $\bz$ generate recourse using a generative model $h$ and neglect control over feature costs (e.g., \citep{pawelczyk2019}). (c): Recourses are found by having the decision maker come up with a causal model between the input features (illustrated by the red directed edges) (e.g., \citep{karimi2020intervention}). (d): Our framework bridges this gap by allowing a generative model $g$ to be influenced by a subset of inputs $\bx_{\mathcal{S}}$. This enables (i) generation of counterfactuals in dense regions of the input space, and (ii) modeling of feature dependencies (iii) without the reliance on causal graphical models.}
\label{fig:spectrum}
\end{figure*}

The fundamental drawbacks of these recourse paradigms
motivate the need for a new recourse framework (see Figure \ref{fig:spectrum}):
(i) The framework should allow recourses to adhere to feature dependencies without relying on causal models. (ii) It should also enable recourses to lie in dense regions of the data distribution.
(iii) Finally, it should ensure that recourses are attainable at low and controllable cost by the individual.
Combining these requirements in one recourse system poses a severe challenge to making algorithmic recourse practicable in the real world.
In this work, we address this critical problem in the face of these three challenges by formulating the problem of algorithmic recourse using a new framework called \texttt{DEAR} (DisEntangling Algorithmic Recourse). 
Our framework exploits a generative model and uses techniques from the disentanglement literature to capture the main practical desiderata (i) -- (iii).
Our key contributions can be summarized as follows:
\begin{itemize}
\item \textbf{Novel recourse framework.} 
Our framework generates recourses by disentangling the latent representation of co-varying features with indirect impact on the recourse from a subset of promising recourse features. 
\item \textbf{Interpretable recourse costs.} 
As a byproduct of our framework, we show that recourse actions can be divided into two types of actions: \emph{direct} and \emph{indirect actions}, 
which can be exploited to lower the cost of algorithmic recourse.
\item \textbf{Constructive theoretical insights.} We develop theoretical expressions for the costs of recourse which guide the design of our generative model and contribute to reliably find low cost algorithmic recourse.
\item \textbf{Extensive experiments.} Our experimental evaluations on real-world data sets demonstrate that \texttt{DEAR} generates significantly less costly and at the same time more realistic recourses than previous manifold based approaches \citep{antoran2020getting,Poyiadzi2020,pawelczyk2019,joshi2019towards}.
\end{itemize}

\section{Related Work}\label{sec:related_work}


Our work builds on a rich literature in the field of algorithmic recourse. We discuss prior works and the connections to this research. 

\textbf{Algorithmic approaches to recourse.}
As discussed earlier, several approaches have been proposed in literature to provide recourse to individuals who have been negatively impacted by model predictions, e.g., \citep{tolomei2017interpretable,laugel2017inverse,Dhurandhar2018,wachter2017counterfactual,ustun2019actionable,joshi2019towards,van2019interpretable,pawelczyk2019,mahajan2019preserving,mothilal2020fat,karimi2019model,rawal2020individualized,karimi2020probabilistic,dandl2020multi,antoran2020getting,spooner2021counterfactual,Poyiadzi2020}. 
These approaches can be roughly categorized
along the following dimensions \cite{verma2020counterfactual}: 
\emph{type of the underlying predictive model} (e.g., tree based \citep{tolomei2017interpretable,lucic2019focus,parmentier2021optimal} vs.\ differentiable classifier \citep{wachter2017counterfactual}), \emph{type of access} they require to the underlying predictive model (e.g., black box \citep{laugel2017inverse,guidotti2019black} vs.\
gradients \citep{antoran2020getting}), whether they encourage \emph{sparsity} in counterfactuals (i.e., only a small number of features should be changed \citep{keane2020good,karimi2019model,schut2021generating}), whether counterfactuals should lie on the \emph{data manifold} \citep{joshi2019towards,pawelczyk2019,mahajan2019preserving,antoran2020getting,Guidotti2019_generative,kenny2020generating,yang2021model}, whether the underlying \emph{causal relationships} should be accounted for when generating counterfactuals \citep{karimi2020intervention,karimi2020probabilistic}, whether the output produced by the method should be \emph{multiple diverse counterfactuals} (e.g., \citep{russell2019efficient,mothilal2020fat}) or a single counterfactual, and whether the underlying \emph{task} is posed as a regression (e.g., \citep{dandl2020multi,spooner2021counterfactual}) or classification problem.

While there have been few recent works that consider input dependencies in algorithmic recourse problems, these works require strong causal assumptions \citep{karimi2020intervention,karimi2020probabilistic}.
For practical use cases, such strong causal assumptions constitute the limiting factor when it comes to the deployment of these models. 
In contrast, our work makes the first attempt at tackling the problem of generating recourses in the presence of feature dependencies while not relying on structural causal models.
 
\textbf{Disentangled representations.}
The techniques that we leverage in this work are inspired by the representation learning literature.
The core principle underlying disentangled representation learning is to learn independent factors of variation that capture well most of the variation underlying the unknown data generating process 
\citep{bengio2013representation}. 
For example, the idea of using disentangled representations has been successfully leveraged to ensure that classifiers are fair while ensuring high classification accuracy downstream \citep{edwards2015censoring,madras2018learning,locatello2019fairness}, to conduct local model audits \citep{marx2019}, or to generate highly realistic data \citep{peebles2020hessian}.
In contrast, our main insight is that disentangled representations can be used to generate recourses in the presence of dependent data by deriving indirect actions from direct actions.

\section{Preliminaries}\label{sec:prelims}


\textbf{Notation.} Before we introduce our framework, we note $\lVert \cdot \rVert$ refers to the 2-norm of a vector, $h(f(\bx))$ denotes the probabilistic output of the trained classifier, where $f: \mathbb{R}^d \to\mathbb{R}$ is a differentiable scoring function (e.g., logit scoring function) and $h: \mathbb{R}\to [0,1]$ is an activation function (e.g., sigmoid) that maps scores to continuous probability outputs. 
We denote the set of outcomes by $y\in \{0,1\}$, where $y=0$ is the undesirable outcome (e.g., loan rejection) and $y=1$ indicates the desirable outcome (e.g., loan approval). 
Moreover, $\hat{y} = \mathbb{I}[h(f(\bx)) > \theta] = \mathbb{I}[f(\bx) > s]$ is the predicted class, where $\mathbb{I}[\cdot]$ denotes the indicator function and $\theta$ is a threshold rule in probability space (e.g., $\theta = 0.5$), with corresponding threshold rule $s$ in scoring space (e.g., $s=0$ when a sigmoid activation is used).

\textbf{The recourse objective.}
Counterfactual explanation methods provide recourses by identifying which attributes to change for reversing an unfavorable model prediction.
We now describe the generic formulation leveraged by several state-of-the-art recourse methods.
The goal is to find a set of actionable changes in order to improve the outcomes of instances $\bx$ which are assigned an undesirable prediction under $f$.
Moreover, one typically defines a cost measure in input space $c: \mathbb{R}^d \times \mathbb{R}^d \xrightarrow{} \mathbb{R}_+$. 
Typical choices are the $\ell_1$ or $\ell_2$ norms.
Then the recourse problem is set up as follows:
\begin{align}
\begin{split}
\bdelta^* = \argmin_{\bdelta} & ~ c(\bx, \xC) 
\text{ s.t. } \xC = \bx + \bdelta, \\
& \xC \in \mathcal{A}_d, ~ f(\xC) = s.
\end{split}
\label{equation:problem_independent}
\end{align}
The objective in eqn.\ \eqref{equation:problem_independent} seeks to minimize the recourse costs $c(\bx, \xC)$ subject to the constraint that the predicted label $\hat{y}$ flips from $0$ (i.e., $f(\xC) < s$) to $1$ (i.e., $f(\xC) \geq s$), and $\mathcal{A}_d$ represents a set of constraints ensuring that only admissible changes are made to the factual input $x$. 
For example, $\mathcal{A}_d$ could specify that no changes to protected attributes such as `sex' can be made.
The assumption underlying \eqref{equation:problem_independent} is that each feature can be independently manipulated regardless of existing feature dependencies.
Under this so-called independently manipulable feature (IMF) assumption, existing popular approaches use gradient based optimization techniques \citep{wachter2017counterfactual,pawelczyk2021connections}, random search \citep{laugel2017inverse}, or integer programming \citep{ustun2019actionable,karimi2019model,rawal2020individualized} to find recourses.


\section{Our Framework: DEAR} \label{section:recourse_independent_mechanism}
The discussion in the previous sections identified three desiderata for a new recourse framework:
\begin{itemize}
\item[(i)] \textbf{Feature dependencies.} The framework should capture feature dependencies while not relying on causal graphical models and structural equations.
\item[(ii)] \textbf{Realistic recourse.} The so identified recourses should lie in dense regions of the input space.
\item[(iii)] \textbf{Low costs.} The framework should allow to find recourses with controllable and low recourse costs.
\end{itemize}

With requirements (i) -- (iii) in mind we present our novel recourse framework, DisEntangling Algorithmic Recourse (\texttt{DEAR}), which satisfies these fundamental requirements.
More specifically:
First, we introduce the generative model required to generate recourses under input dependencies that lie in dense regions of the input space.
Second, using our model we then show that disentangled representations need to be learned to yield accurate recourse cost estimates.
Third, we present our objective function to generate recourses under input dependencies and suggest a constructive explanation for why our framework finds recourses more reliably than existing manifold-based approaches. 
Finally, we provide a detailed discussion on how to operationalize and optimize our objective effectively.

\subsection{The Generative Model}
On a high level, our framework consists of separating the latent code of a generative model into 1) observable features $\bx_{\mathcal{S}}$ -- that we wish to perform direct recourse actions on -- and 2) latent space features $\bv$ that have been trained to become disentangled of the observable features. 
A direct recourse action has two effects: a direct effect on the input features that have to be changed, and an indirect effect on other, dependent features. 
The strength of the indirect effect is then determined by a generative model (see Figure \ref{fig:aae}). 
To formalize this intuition, let the input $\bx$ be produced by the following generative model:
\begin{align}
\begin{split}
\bx & = [\bx_{\mathcal{S}}, \bx_{S^c}] \\
&= [ g_{\bx_{\mathcal{S}}}(\bv,\bx_{\mathcal{S}}),  g_{\bx_{S^c}} (\bv,\bx_{\mathcal{S}})]
= g(\bv, \bx_{\mathcal{S}}),
\label{eq:generative_model_indepence}
\end{split}
\end{align}
where $g: \mathbb{R}^k \to \mathbb{R}^d$, $\bv \in \mathbb{R}^{k-|\mathcal{S}|}$ refers to the latent code and $\bx_{\mathcal{S}}$ corresponds to a subset of the input features where $S \subset  \{1, \dots, d\}$, and the complement set is $S^c = \{1, \dots, d\} ~ \backslash ~ S$. 


\subsection{Disentangled Representations Promote Low Costs}
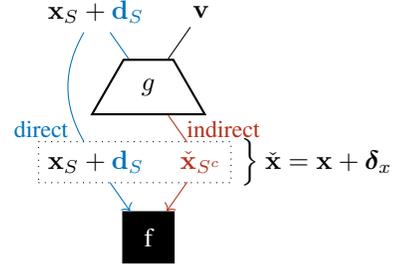
\begin{figure}[tb]
\centering
\usetikzlibrary{shapes.geometric}
\usetikzlibrary{fit}
\begin{tikzpicture}[square/.style={regular polygon,regular polygon sides=4}]

  \node (latent1) at(+0.7,1.0) {$\bv$};
  \node (latent2) at(-0.7,1.0) {$\bx_S+\textcolor{RoyalBlue}{\bd_{S}}$};
  
  \node (dec) [trapezium, trapezium angle=60, minimum width=15mm, draw, thick] at(0,0) {$g$};
 
  \node (out1) at (+0.7,-1.0) {$\textcolor{BrickRed}{\check{\bx}_{S^c}}$};
  \node (out2) at (-0.7,-1.0) {$\bx_S+ \textcolor{RoyalBlue}{\bd_{S}}$};
  
  \node[draw, dotted, inner sep=0pt, fit=(out1) (out2), label=right: {$ \Big\}$  $\xC = \bx + \bdelta_{x} $}] {} ;

  \node (f) [square, fill=black, draw, thick] at (0,-2) {\textcolor{white}{f}};
  
 
 \draw (out1) edge[BrickRed,-] node [right] (TextNode1) {\rotatebox{0}{\small \text{indirect}}}  (dec);
 \draw[RoyalBlue] (latent2) to  [bend right=30] node [pos=0.89, left] (TextNode1) {\rotatebox{0}{\small \text{direct}}} (out2);

 \draw (latent1) -- (dec);
 \draw (latent2) edge[RoyalBlue,-] (dec);

 \draw (out1) edge[BrickRed, ->] (f);
 \draw (out2) edge[RoyalBlue,->] (f);

\end{tikzpicture}
\caption{Finding recourse for input $\bx$ with \texttt{DEAR}.
For an input $\bx$ with $f(\bx) < s$, we encode $[\bx_{\mathcal{S}}, \bv]{=}e(\bx)$. Then, we find direct actions $\textcolor{RoyalBlue}{\bd_{\mathcal{S}}}$, which generate recourse, i.e.\ we find a $\xC$ such that $f(\xC) = s$, where $\xC = [\textcolor{RoyalBlue}{\bd_{\mathcal{S}} } + \bx_{\mathcal{S}},\textcolor{BrickRed}{g_{\bx_{S^c}}}(\bv, \textcolor{RoyalBlue}{\bd_{\mathcal{S}}} + \bx_{\mathcal{S}})]$. The direct action \textcolor{RoyalBlue}{$\bd_{\mathcal{S}}$} has two effects: 1) it changes the features in $\mathcal{S}$ directly (i.e., $\textcolor{RoyalBlue}{\bd_{\mathcal{S}} } + \bx_{\mathcal{S}}$ ), and 2) it changes the features in $\mathcal{S}^c$ indirectly (i.e., $\textcolor{BrickRed}{\check{\bx}_{S^c}} = \textcolor{BrickRed}{g_{\bx_{S^c}}}(\bv, \textcolor{RoyalBlue}{\bd_{\mathcal{S}}} + \bx_{\mathcal{S}}))$. The strength of the indirect change $\textcolor{BrickRed}{\check{\bx}_{S^c}}$ is determined by the \text{elasticity of $g_{\bx_{S^c}}$ w.r.t.\  $\bx_{\mathcal{S}}$}.} 
\label{fig:aae}
\end{figure}

Since one of the key considerations in recourse literature are recourse costs we use our generative model from eqn.\ \eqref{eq:generative_model_indepence} and analyze the recourse cost estimates under this model. 
Using the following Proposition, we obtain an intuition on how the generative model required for our framework has to be trained to yield low recourse costs.
\begin{proposition}[Recourse costs]
Under the generative model in \eqref{eq:generative_model_indepence}, the cost of recourse $\lVert \bdelta_x \rVert^2 = \lVert \bx - \xC \rVert^2$ is given by:
\begin{align*}
\begin{split}
 \lVert \bdelta_x \rVert^2
 & \approx \underbrace{\bd^\top_{\mathcal{S}} \big( {\bJ_{\bx_{\mathcal{S}}}^{(\bx_{\mathcal{S}})}}^\top {\bJ_{\bx_{\mathcal{S}}}^{(\bx_{\mathcal{S}})}} \big) \bd_{\mathcal{S}}}_{\textcolor{RoyalBlue}{\text{Direct Costs}}} + \underbrace{\bd^\top_{\mathcal{S}} \big({\bJ_{\bx_{\mathcal{S}}}^{(\bx_{S^c})}}^\top \bJ_{\bx_{\mathcal{S}}}^{(\bx_{S^c})}\big) \bd_{\mathcal{S}}}_{\textcolor{BrickRed}{\text{Indirect Costs}}}, 
\end{split}
\end{align*}
\begin{align*}
\begin{split}
where: \\
{\bJ_{\bx_{\mathcal{S}}}^{(\bx_{\mathcal{S}})}} &= \underset{\textcolor{RoyalBlue}{\text{Entanglement costs}}}{\underbrace{\frac{\partial g_{\bx_{\mathcal{S}}}(\bv, \bx_{\mathcal{S}})}{\partial \bv} \frac{\partial \bv}{\partial \bx_{\mathcal{S}}}}} + \underset{\textcolor{RoyalBlue}{\text{Identity Mapping}}}{\underbrace{\frac{\partial g_{\bx_{\mathcal{S}}}(\bv, \bx_{\mathcal{S}})}{\partial\bx_{\mathcal{S}}}}} \\
{\bJ_{\bx_{\mathcal{S}}}^{(\bx_{S^c})}} & = \underset{\textcolor{BrickRed}{\text{Entanglement costs}}}{\underbrace{\frac{\partial g_{\bx_{S^c}}(\bv, \bx_{\mathcal{S}})}{\partial \bv} \frac{\partial \bv}{\partial \bx_{\mathcal{S}}}}} + \underset{\textcolor{BrickRed}{\text{Elasticity of $g_{\bx_{S^c}}$ w.r.t  $\bx_{\mathcal{S}}$}}}{\underbrace{\frac{\partial g_{\bx_{S^c}}(\bv, \bx_{\mathcal{S}})}{\partial \bx_{\mathcal{S}}}}}.
\end{split}
\end{align*}
\label{proposition:recourse_costs_ae}
\end{proposition}
The result of Proposition \ref{proposition:recourse_costs_ae} provides constructive insights for the implementation of the generative model $g$.
It says that we can control the recourse costs using the actions $\bd_{\mathcal{S}}$.
First, the result reveals that the costs have to be partitioned into \textcolor{RoyalBlue}{direct} and \textcolor{BrickRed}{indirect} costs. 
The direct costs correspond to the costs that one would have obtained from algorithms that use the IMF assumption when searching for recourses (e.g., \citep{wachter2017counterfactual,pawelczyk2021connections,ustun2019actionable}). 
The indirect costs are due to feature dependencies of $\bx_{\mathcal{S}}$ with $\bx_{S^c}$. 
If $\bx_{\mathcal{S}}$ is independent of $\bx_{S^c}$ (i.e., the elasticity of $g_{\bx_{S^c}}$ w.r.t  $\bx_{\mathcal{S}}$ is $\mathbf{0}$), then a change in $\bx_{\mathcal{S}}$ will not alter $\bx_{S^c}$ and the only cost remaining is the direct cost (we refer to Figure \ref{fig:aae} for a schematic overview of the mechanism). 
Second, we observe that the costs can be \emph{inflated}, if the latent space variables $\bv$ depend on $\bx_{\mathcal{S}}$. 
This is expressed through the \emph{entanglement cost} terms in Proposition \ref{proposition:recourse_costs_ae} and suggests that the model $g$ should be trained such that $\bx_{\mathcal{S}} \upmodels \bv$ to keep the recourse costs low. 

\subsection{Our Recourse Objective} \label{section:recourse_model}
So far the predictive model $f$ has played no role in our considerations.
Now, we introduce the predictive model to rewrite the recourse problem from \eqref{equation:problem_independent} as follows:
\begin{align}
\begin{split}
\bd_{\mathcal{S}}^* = & \argmin_{\bd_{\mathcal{S}}} c(\bx, \xC )
\text{ s.t. } \xC = g(\bv ,\bx_{\mathcal{S}} + \bd_{\mathcal{S}}),\\
\xC &\in \mathcal{A}_d, ~ \bx_{\mathcal{S}} \upmodels \bv, ~
f(\xC) = s,
\end{split}
\label{equation:problem_dependent_partition}
\end{align}
where we have used the insight that $\bx_{\mathcal{S}} \upmodels \bv$ derived from Proposition \ref{proposition:recourse_costs_ae}.
Relative to the objective from eqn.\ \eqref{equation:problem_independent}, our objective in eqn.\ \eqref{equation:problem_dependent_partition} uses our generative model to capture input dependencies. 
Instead of finding recourse actions across the whole input space, we find recourse actions for the inputs in $\mathcal{S}$.
We make recommendation on the choice of $\mathcal{S}$ in Section \ref{sec:aligned_gen_model}.
Our reformulation has several advantages compared to existing recourse methods from the literature:
i) relative to manifold-based recourse methods \citep{joshi2019towards,pawelczyk2019,antoran2020getting} the actions are applied to input space variables as opposed to latent space variables, and thus they are inherently interpretable;
ii) relative to manifold-based recourse methods and recourse methods which use the IMF assumption \citep{wachter2017counterfactual,laugel2017inverse}, we can sharply separate the direct effect, which $\bd_{\mathcal{S}}$ has on $\xC$ via $\bx_{\mathcal{S}}$, from its indirect effect, which $\bd_{\mathcal{S}}$ has on $\xC$ determined by the generative model when it is dependent of $\bx_{S^c}$ (recall Proposition \ref{proposition:recourse_costs_ae});
iii) relative to causal recourse methods \citep{karimi2020intervention,karimi2020probabilistic}, we neither assumed causal graphical models nor did we assume structural equation models to incorporate input dependencies. 

\subsection{Aligned Generative Models Promote Finding Recourses Reliably}
\label{sec:aligned_gen_model}
So far we have learned how disentangled represenations help reduce the recourse costs. 
Related work \citep{antoran2020getting} has reported that manifold-based methods, which search for recourse in latent space (e.g., \cite{pawelczyk2019,joshi2019towards,antoran2020getting}), sometimes get stuck before they find a recourse. 
In this section, we develop a theoretical expression that will inform the choice of the set $\mathcal{S}$, and we will see that choosing this set appropriately promotes finding recourses more reliably. 
To this end, we derive an approximate closed-form solution for the objective in \eqref{equation:problem_dependent_partition} which uses our insights from Proposition \ref{proposition:recourse_costs_ae} (i.e., $\bv \upmodels \bx_{\mathcal{S}}$). 
\begin{proposition}[Direct action]
Given $\bv \upmodels \bx_{\mathcal{S}}$ and $c=\lVert \bx - \xC \rVert^2$, a first-order approximation $\tilde{\bd}_{\mathcal{S}}^*$ to the optimal direct action $\bd_{\mathcal{S}}^*$ from the objective in \eqref{equation:problem_dependent_partition} is given by:
\begin{align}
\tilde{\bd}_{\mathcal{S}}^* = \frac{m}{\lambda + \lVert \bw \rVert_2^2} \cdot \bw,
\label{equation:closed_form_dear}
\end{align}
where $\left.\bY_{\bx_{\mathcal{S}}}^{(\bx)}{:=} \frac{\partial g(\bv, \bx_{\mathcal{S}})}{\partial \bx_{\mathcal{S}} }\right\vert_{\bv=\bv, \bx_{\mathcal{S}}=\bx_{\mathcal{S}}}$, $s$ is the target score,
$m=s{-}f(\bx)$, $\bw{=} {\bY_{\bx_{\mathcal{S}}}^{(\bx)}}^\top \nabla f(\bx)$ and $\lambda$ is the trade-off parameter.
\label{proposition:dear}
\end{proposition}

\begin{corollary}[Recourse action]
Under the same conditions stated in Proposition \ref{proposition:dear}, a first-order approximation to the recourse action is given by:
\begin{align}
\bdelta^*_x \approx {\bY_{\bx_{\mathcal{S}}}^{(\bx)}} \tilde{\bd}_{\mathcal{S}}^* = \frac{m}{\lambda + \lVert \bw \rVert_2^2} \cdot {\bY_{\bx_{\mathcal{S}}}^{(\bx)}} \bw.
\label{equation:closed_form_dear_corrolary}
\end{align}
\end{corollary}


The above result is intuitive. The optimal recourse action $\tilde{\bd}_{\mathcal{S}}^*$ applied to the inputs $\bx_{\mathcal{S}}$ is being transformed by the generator Jacobian ${\bY_{\bx_{\mathcal{S}}}^{(\bx)}}$ to yield the optimal action in input space $\bdelta^*_x$. 
The generator Jacobian, in turn, measures the influence that the features in $\mathcal{S}$ have on the input $\bx$. 

We suggest to use \emph{singletons} for the set of variables $\mathcal{S}$ where the direct action should be performed on as these are easiest to interpret (see Appendix \ref{appendix:case_study}) and reliably lead to low cost recourses.
Using singletons the insight from Proposition \ref{proposition:dear} becomes more clear.
Then $\bw$ from  eqn.\ \eqref{equation:closed_form_dear} is a scalar.
Therefore, to make most progress towards the desired outcome at first order, the generator Jaocobian $\bY_{\bx_{\mathcal{S}}}^{(\bx)}$ should be aligned with the model gradient $\nabla f(\bx)$, i.e., the two vectors should have a high similarity in the dot-product sense. 
To see this, consider a first-order approximation of $f(\bx + \bdelta^*_x) \approx f(\bx) + \nabla f(\bx)^\top \bdelta^*_x  = f(\bx) + \frac{m \cdot w}{\lambda + \lVert \bw \rVert_2^2} \cdot  \nabla f(\bx)^\top \bY_{\bx_{\mathcal{S}}}^{(\bx)}$.
To push the score $f(\bx)$ towards the target score $s\geq 0$, $\mathcal{S}$ should be chosen such that the dot product $\nabla f(\bx)^\top \bY_{\bx_{\mathcal{S}}}^{(\bx)}$ is high.


\subsection{Optimizing our Objective} \label{section:marie_algorithm}
Motivated by the insights from Propositions \ref{proposition:recourse_costs_ae} and \ref{proposition:dear}, we present an algorithmic procedure to compute minimal cost recourses under feature dependencies using a penalty term during autoencoder training that encourages disentanglement of $\bx_{\mathcal{S}}$ and $\bv$ in order to keep the \emph{entanglement costs} low.
In summary, \texttt{DEAR} requires two steps: first, we need to obtain a latent space representation $\bv$, which is independent of $\bx_{\mathcal{S}}$. 
Second, we require an optimization procedure to identify the nearest counterfactual.

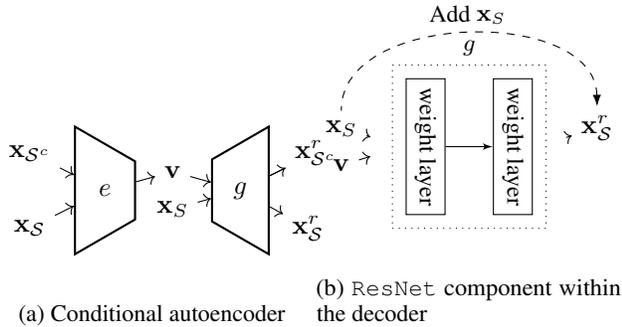
\begin{figure}[htb]
\raggedleft
\begin{subfigure}[b]{0.49\columnwidth}
\centering
\usetikzlibrary{shapes.geometric}
\begin{tikzpicture}[square/.style={regular polygon,regular polygon sides=4}]

  \node (in1) at(-1.9,0.5) {$\bx_{\mathcal{S}^c}$};
  \node (in2) at(-1.9,-0.5) {$\bx_{\mathcal{S}}$};
  
  \node (out1) at(+1.9,0.5) {$\bx^r_{\mathcal{S}^c}$};
  \node (out2) at(+1.8,-0.5) {$\bx^r_{\mathcal{S}}$};
  
  \node (latent1) at(+0,0.25) {$\bv$};
  \node (latent2) at(+0,-0.25) {$\bx_S$};
  
  \node (enc) [trapezium, trapezium angle=60, minimum width=17mm, draw, thick, rotate=270] at (-0.9,0) {\rotatebox{90}{$e$}};
  
  \node (dec) [trapezium, trapezium angle=60, minimum width=17mm, draw, thick, rotate=90] at(0.9,0) {\rotatebox{270}{$g$}};

 \draw (in1) edge[->] (enc);
 \draw (in2) edge[->] (enc);
  
 \draw (out1) edge[-]  (dec);

 \draw (enc) edge[->] (latent1);

 \draw (latent1) edge[->] (dec);
 \draw (latent2) edge[->] (dec);
 
 \draw (dec) edge[->] (out1);
  \draw (dec) edge[->] (out2);
 

\end{tikzpicture}
\caption{Conditional autoencoder}  
\label{fig:c_ae}
\end{subfigure} 
\hfill
\begin{subfigure}[b]{0.49\columnwidth}
\centering 
\usetikzlibrary{shapes.geometric}
\begin{tikzpicture}[square/.style={regular polygon,regular polygon sides=4}]
\tikzstyle{arrow} = [draw, -latex']

 \node (in1) at(-1.7,0.25) {$\bx_S$};
 \node (in2) at(-1.7,-0.25) {$\bv$};
  
 \node (out1) at(+1.7,0.25) {$\bx^r_{\mathcal{S}}$};

 \node (dec1) [rectangle, minimum width=16mm, draw, thin, rotate=270] at (-0.58,0) {\small weight layer};
  
 \node (dec2) [rectangle, minimum width=16mm, draw, thin, rotate=270] at (+0.58,0) {\small weight layer};

 \node[draw, dotted, inner sep=5pt, fit=(dec1) (dec2), label=above: \small $g$] {} ;
  
 \draw (in1) edge[->] (enc);
 \draw (in2) edge[->] (enc);

 
\path [arrow] (dec1) -- node (a) [above] {\small } (dec2) ;

 \draw (dec) edge[->] (out1);

 
 \draw [{Latex[length=1.5mm]}-, dashed] (out1) to [bend right=90] node [above, sloped] (TextNode1) {\small Add $\bx_S$} (in1);

\end{tikzpicture}
\caption{\texttt{ResNet} component within the decoder}
\end{subfigure}
\caption{\texttt{DEAR}'s autoencoder architecture.
(a): The conditional autoencoder architecture is trained subject to the Hessian penalty described in Section \ref{section:marie_algorithm}. (b): We achieve the identity mapping between $\bx_{\mathcal{S}}$ and the reconstructed $\bx^r_{\mathcal{S}}$ using a \texttt{ResNet} component \citep{he2016deep,he2016identity}: we add $\bx_{\mathcal{S}}$ to $\bx^r_{\mathcal{S}}$  before passing the arguments to the loss $\mathcal{L}_R$ from eqn.\ \eqref{eq:reconstruction_loss}.}
\label{figure:architecture}
\end{figure}

\textbf{Step 1: Training the Generative Model.}
The main idea is to train a generative model, in which $\bx_{\mathcal{S}}$ is independent of the latent variable $\bv$, while providing high-quality reconstruction of the input $\bx$.
Thus, the training loss for the generative model consists of two components. 
First, it consists of both an encoder network $e$ and decoder network $g$, for which the reconstruction loss,
\begin{align}
\mathcal{L}_R(g,e; \bx_{\mathcal{S}}) =\lVert g( e(\bx), \bx_{\mathcal{S}}) - \bx \rVert_2^2,
\label{eq:reconstruction_loss}
\end{align}
guides both networks towards a good reconstruction of $\bx$.
Second, we want to drive the \emph{entanglement costs} to 0, for which we need the decoder $g$ to be disentangled with respect to the latent space, i.e., each component of $\bz = [\bv, \bx_{\mathcal{S}}]$ should ideally control a single factor of variation in the output of $g$. 
To formalize this intuition, recall that $g(\bx_{\mathcal{S}}, \bv) = \bx \in \mathbb{R}^d$, where each output $g_j=x_j$ for $1 \leq j \leq d$ has its own $|\bx_{\mathcal{S}}| \times |\bv|$ Hessian matrix $\bH^{(j)}$. 
We refer to the collections of the $d$ Hessian matrices as $\bH$.
Thus, the second loss we seek to minimize is given by:
\begin{align}
    \mathcal{L}_{\bH}(g; \bx_{\mathcal{S}}) = \sum_{j=1}^d \bigg( \sum_{k=1}^{|\bv|} \sum_{l=1}^{|\bx_{\mathcal{S}}|} H_{kl}^{(j)} \bigg),
\label{eq:hessian_penalty}
\end{align}
which is also known as the Hessian penalty \citep{peebles2020hessian}.
We illustrate the intuition of this objective on the $j$-th output $x_j$: we regularize the Hessian matrix $\bH^{(j)} = \frac{\partial}{\partial \bv} \frac{\partial g_j}{\partial \bx_{\mathcal{S}}}$ and encourage its off-diagonal terms to become $0$. 
Driving the off-diagonal terms to $0$ implies that $\frac{\partial g_j}{\partial \bx_{\mathcal{S}}}$ is not a function of $\bv$ and thus $\bv$ plays no role for the output of $g_j$ when searching for minimum cost actions using $\bx_{\mathcal{S}}$.
We use the Hessian penalty from \citep{peebles2020hessian} in our implementation.
Finally, Proposition \ref{proposition:recourse_costs_ae} requires the identity mapping between the latent space $\bx_{\mathcal{S}}$ and the reconstructed $\bx^r_{\mathcal{S}}$.
We encourage our generator $g$ to learn this mapping by using a \texttt{ResNet} architecture \citep{he2016deep,he2016identity} as shown in Figure \ref{figure:architecture}.

\textbf{Step 2: Finding Minimal Cost Actions $\bd_{\mathcal{S}}$.} Given our trained generative model from step 1, we rewrite the problem in eqn.\ \eqref{equation:problem_dependent_partition} using a Lagrangian with trade-off parameter $\lambda$. 
For a given encoded input instance $e(\bx) = [\bv, \bx_{\mathcal{S}}]$, our objective function reads:
\begin{align}
\begin{split}
\bd^*_{\mathcal{S}} &= \argmin_{\bd_{\mathcal{S}}, ~ \xC \in \mathcal{A}_d} \mathcal{L} \\
& = \argmin_{\bd_{\mathcal{S}}, ~ \xC \in \mathcal{A}_d} \ell\big(f(\xC), s \big) + \lambda \lVert \bx - \xC) \lVert_1,
\end{split}
\label{equation:surrogate_loss_dear}
\end{align}
where $\xC = g(\bv, \bx_{\mathcal{S}}+\bd_{\mathcal{S}})$ is a potential counterfactual in input space, $\ell(\cdot, \cdot)$ denotes the MSE loss, and $s\geq0$ is the target score in logit space. 
The term on the right side encourages the counterfactual $g(\bv, \bx_{\mathcal{S}}+\bd_{\mathcal{S}}) = \xC$ to be close to the given input point $\bx$, while the left hand side encourages the predictions to be pushed from the factual output $f(\bx)$ towards s. 
We do gradient descent iteratively on the loss function in eqn.\ \eqref{equation:surrogate_loss_dear} until the class label changes from $y=0$ to $y=1$.
Algorithm \ref{algorithm:dear_differentiable} summarizes our optimization procedure.
Finally, in Appendix \ref{Appendix:architecture_params} we further discuss how monotonicity constraints and how categorical variables are included in our objective.
\begin{algorithm}[tb]
   \caption{\texttt{DEAR}}
   \label{alg:example}
\begin{algorithmic}
   \STATE {\bfseries Input:} $f$, $\bx$ s.t.\ $f(\bx){<}0$, $g$, $e$, $\lambda> 0$, Learning rate: $\alpha>0$, $s\geq 0$, $\mathcal{S}$
   \STATE {\bfseries Initialize:} $\bd_{\mathcal{S}} = \mathbf{0}$, $ e(\bx)=[\bv, \bx_{\mathcal{S}}]$, $\xC = g(\bv, \bx_{\mathcal{S}} + \bd_{\mathcal{S}})$
   \WHILE{$f(\xC)< s$} 
   \STATE $\bd_{\mathcal{S}} =  \bd_{\mathcal{S}} - \alpha \cdot \nabla_{\bd_{\mathcal{S}}} \mathcal{L}(\xC;f, s, \lambda)$ \COMMENT{Optimize
   \eqref{equation:surrogate_loss_dear}}
   \STATE $\xC = g(\bv, \bx_{\mathcal{S}}+\bd_{\mathcal{S}})$
   \ENDWHILE ~ \COMMENT{Class changed, i.e., $f(\xC) \geq s$}
   \STATE {\bfseries Return:} $\xC^* = \xC$ 
\end{algorithmic}
\label{algorithm:dear_differentiable}
\end{algorithm}

\section{Experimental Results} \label{section:experiments}

\begin{figure*}[tb]
\centering
\begin{subfigure}[b]{\textwidth}
\centering
\includegraphics[width=\textwidth]{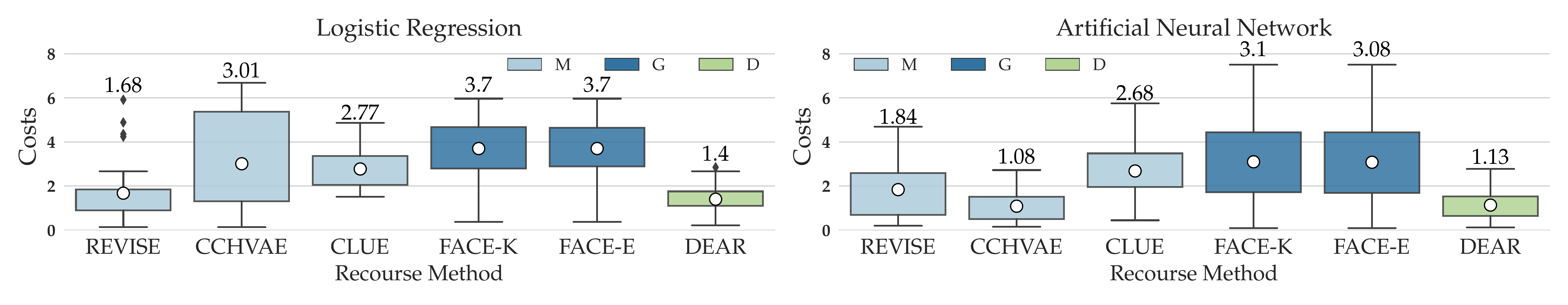}
\caption{Adult}
\label{fig:adult_cost_comparison}
\end{subfigure}
\vfill
\begin{subfigure}[b]{\textwidth}
\centering
\includegraphics[width=\textwidth]{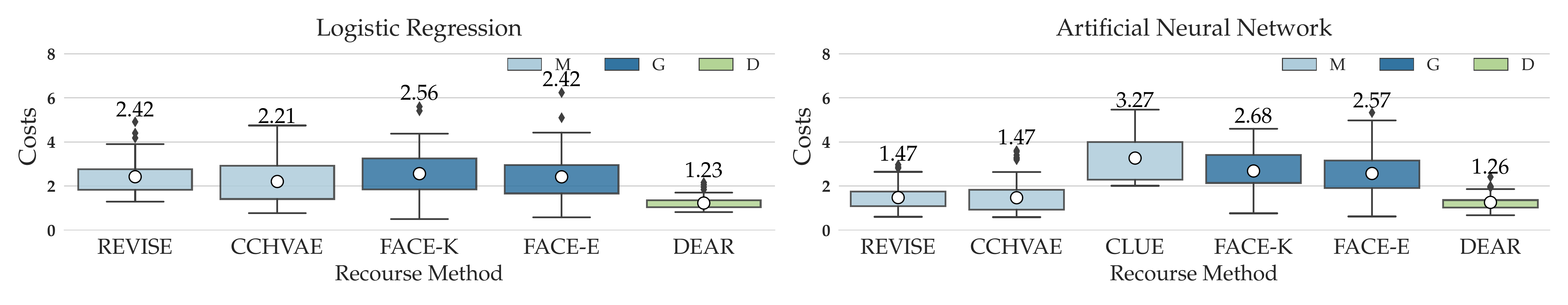}
\caption{COMPAS}
\label{fig:compas_cost_comparison}
\end{subfigure}
\caption{Measuring the cost of recourse ($\ell_1$) across recourse methods, demonstrating that \texttt{DEAR} reduces recourse costs by up to 49\% (bottom left).
We use boxplots to show the distribution of recourse costs across all individuals from the test set who require algorithmic recourse. 
The numbers above the maximum values correspond to the white dots (= median recourse costs). 
The color of the boxplots represent the type of the recourse method: The methods 
with `M' purely focus on the data manifold constraints, and methods with `G' use a graphical model to generate recourses through dense paths. `D' refers to our method which takes both manifold and input dependencies into account.
Results for GMC are relegated to Appendix \ref{appendix:case_study}.
}
\label{fig:cost_results}
\end{figure*}

In this Section, we conduct extensive quantitative and qualitative evaluations to analyze \texttt{DEAR}'s performance using our conceptual insights from the previous section.
\emph{Quantitatively}, we conduct a baseline comparison contrasting our framework \texttt{DEAR} with state-of-the-art recourse methods, which use generative models, using common evaluation measures \citep{pawelczyk2021carla} from the recourse literature such as recourse costs and reliability measures.
\emph{Qualitatively}, we consider three aspects: (i) the \emph{entanglement costs}, (ii) the structure of the \emph{cost splits} (i.e., direct vs.\ indirect costs) and (iii) a \emph{case study} to showcase the advantages of our new framework which we relegated to Appendix \ref{appendix:case_study}. 

\definecolor{Gray}{gray}{0.9}
\begin{table*}[tb]
\centering
\resizebox{\textwidth}{!}{%
\begin{tabular}{clcccccccccccccccccc}
\toprule
&&
\multicolumn{6}{c}{Adult} & \multicolumn{6}{c}{COMPAS}  & \multicolumn{6}{c}{GMC}  \\
\cmidrule(lr){3-8} \cmidrule(lr){9-14} \cmidrule(lr){15-20}
&& \multicolumn{3}{c}{LR} & \multicolumn{3}{c}{ANN} & \multicolumn{3}{c}{LR} & \multicolumn{3}{c}{ANN} & \multicolumn{3}{c}{LR} & \multicolumn{3}{c}{ANN} \\
\cmidrule(lr){3-5} \cmidrule(lr){6-8} \cmidrule(lr){9-11} \cmidrule(lr){12-14} \cmidrule(lr){15-17} \cmidrule(lr){18-20}
\multicolumn{2}{c}{Method} & SR ($\uparrow$) & CV ($\downarrow$) & YNN ($\uparrow$) & SR ($\uparrow$) & CV ($\downarrow$) & YNN ($\uparrow$) & SR ($\uparrow$) & CV ($\downarrow$) & YNN ($\uparrow$) & SR ($\uparrow$) & CV ($\downarrow$) & YNN ($\uparrow$) & SR ($\uparrow$) & CV ($\downarrow$) & YNN ($\uparrow$) & SR ($\uparrow$) & CV ($\downarrow$) & YNN ($\uparrow$) \\
\cmidrule(lr){1-2} 
\cmidrule(lr){3-5} \cmidrule(lr){6-8} \cmidrule(lr){9-11} \cmidrule(lr){12-14}
\cmidrule(lr){15-17} \cmidrule(lr){18-20}
\multirow{4}{*}{M} &\texttt{REVISE} & 0.35	& 0.00$^*$	& 0.37	&\textbf{1.00}	& 0.12	&\textbf{0.72}	&0.63	& 0.11	&\textbf{1.00}	& 0.99	& 0.12 	& 0.98	&0.99	& NA	&\textbf{\textbf{1.00}}	&\textbf{\textbf{1.00}}	& NA	&0.95 \\
&\texttt{CCHVAE} &0.54	& 0.17	&0.53	&\textbf{1.00}	& 0.07	&0.61	&\textbf{1.00}	& 0.32	&\textbf{1.00}	&\textbf{1.00}	& 0.17	&0.96	&\textbf{1.00}	& NA	&0.24	&\textbf{1.00}	& NA	&0.80 \\
&\texttt{CLUE} & 0.38$^*$ & 0.00$^*$ & 1.00$^*$ & \textbf{1.00} & 0.00 & 0.25 & 0.00 & -- & -- & 0.21 & 0.00 & 1.00 & \textbf{1.00} & NA & \textbf{1.00} & \textbf{1.00} & NA & 0.92  \\
\midrule
\multirow{2}{*}{G} &\texttt{FACE-K} & 0.99	& 0.36	&0.71	&\textbf{1.00}	& 0.29	&0.57	&0.99	& 0.38	&\textbf{1.00}	&0.60	& 0.40	&\textbf{\textbf{1.00}}	&\textbf{\textbf{1.00}}	& NA	& 0.95	&\textbf{1.00}	& NA	& 0.96 \\
&\texttt{FACE-E} &0.74	& 0.38 &0.70	&0.99	& 0.30	&0.58	&0.99	& 0.41	&\textbf{1.00}	&0.39	& 0.40	&\textbf{1.00}	&\textbf{1.00}	& NA	&0.94	&\textbf{1.00}	& NA	&\textbf{0.97} \\
\rowcolor{Gray}
\midrule
D &\texttt{DEAR} & \textbf{1.00}	& \textbf{0.00} 	& \textbf{0.84}	& \textbf{1.00}	& \textbf{0.00}	& 0.70	&\textbf{1.00}	& \textbf{0.00}	&\textbf{1.00}	&\textbf{1.00}	& 0.01	&\textbf{1.00}	&\textbf{1.00}	& NA	&0.91	&\textbf{1.00}	& NA	&0.94
   \\
\bottomrule
\end{tabular}%
}
\caption{Measuring the reliability of algorithmic recourse for the ANN and LR models on all data sets. The \emph{success rate} (SR), \emph{constraint violation} (CV) and \emph{$y$-nearest neighbors} (YNN) measures are described in Section \ref{section:experiments}. For GMC, there were no immutable features and therefore we are reporting NA. 
Our method (i.e., \texttt{DEAR}) usually performs on par or better relative to other recourse methods. $^*$: Methods with success rates below $50\%$ are excluded from the evaluation.}
\label{tab:reliability}
\end{table*}

\begin{figure*}[tb]
\centering
\begin{subfigure}[b]{0.32\textwidth}
\centering
\includegraphics[width=\textwidth]{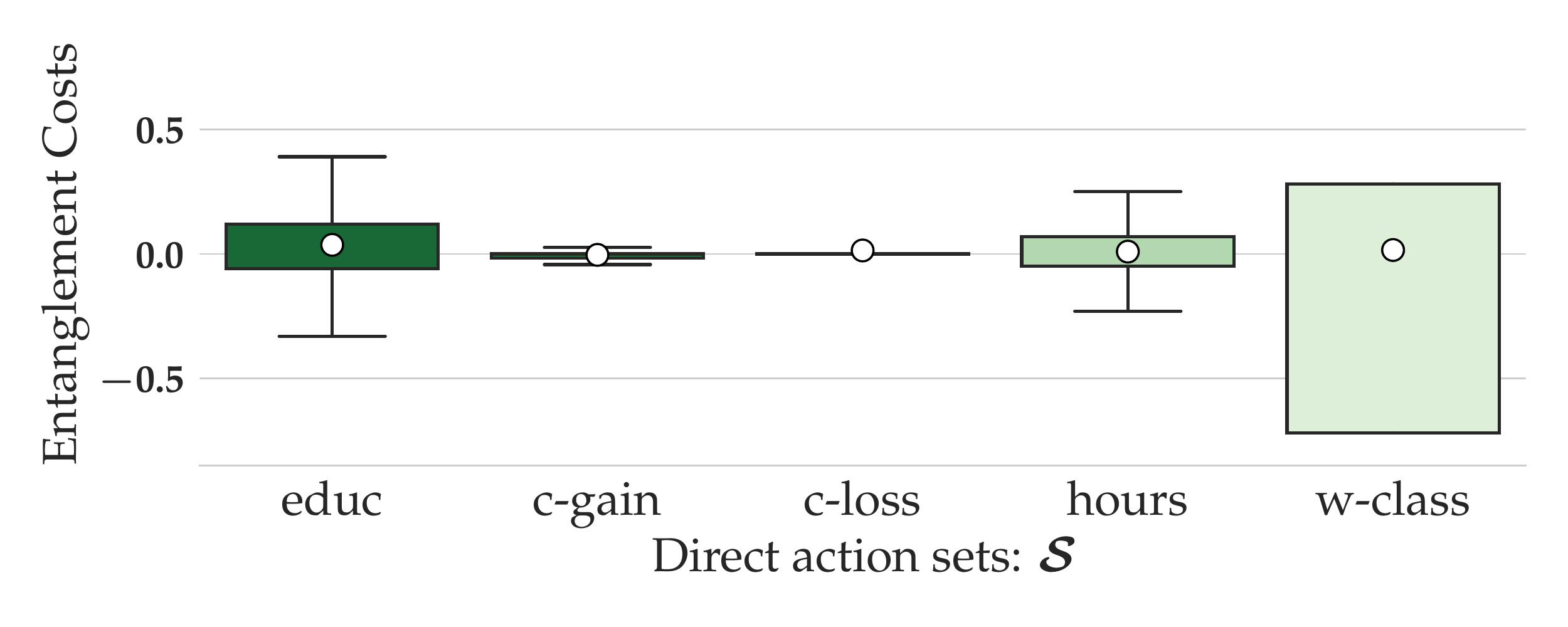}
\caption{Adult}
\label{fig:error_adult}
\end{subfigure}
\hfill
\begin{subfigure}[b]{0.32\textwidth}
\centering
\includegraphics[width=\textwidth]{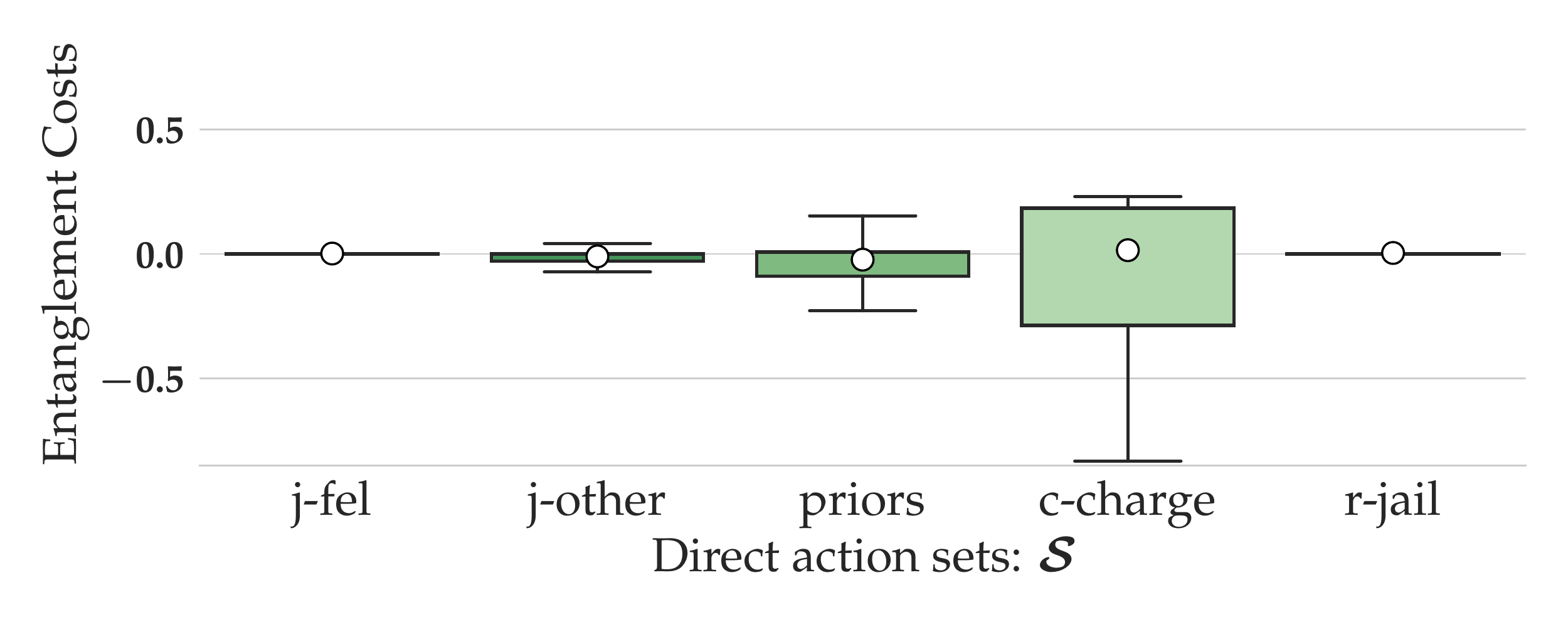}
\caption{COMPAS}
\label{fig:error_compas}
\end{subfigure}
\hfill
\begin{subfigure}[b]{0.32\textwidth}
\centering
\includegraphics[width=\textwidth]{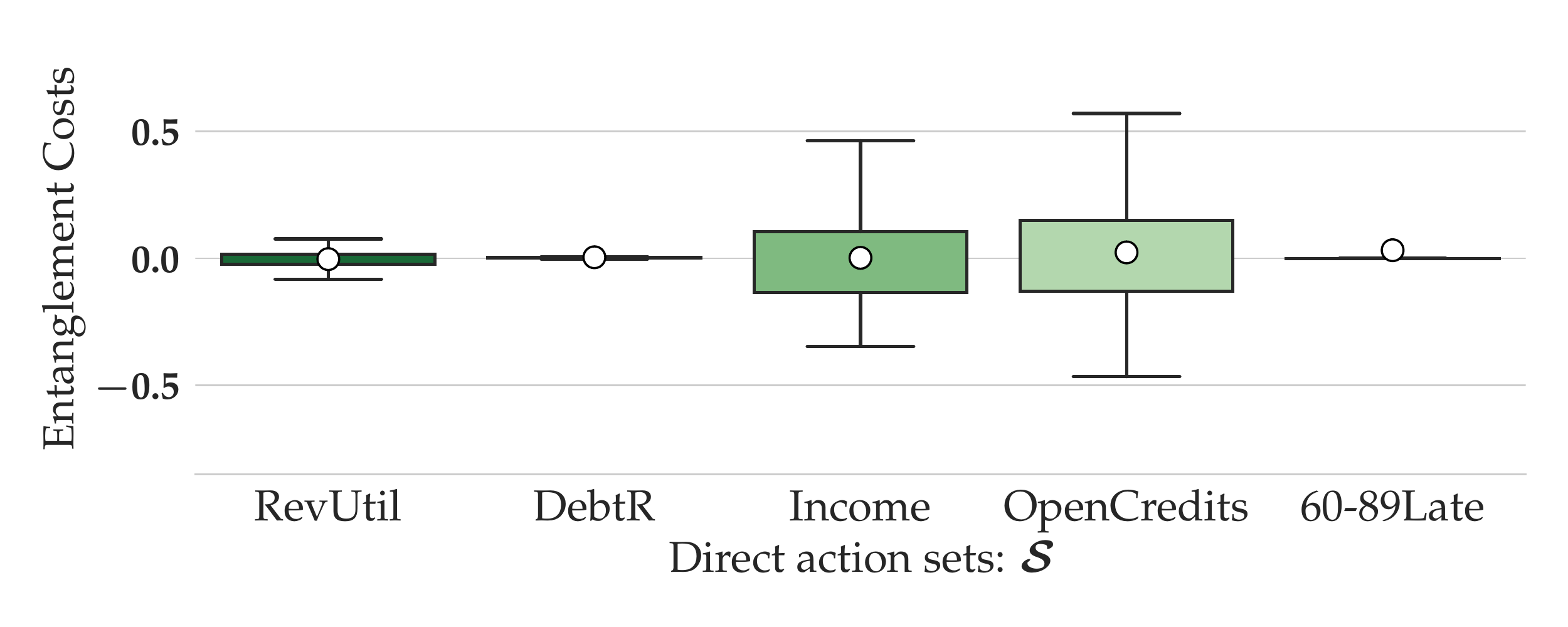}
\caption{GMC}
\label{fig:error_gmc}
\end{subfigure}
\caption{Evaluating \texttt{DEAR}'s entanglement costs on all data sets. 
At the end of autoencoder training, we compute the Hessians' off-diagonal elements of the decoder and average them (see Section \ref{section:experimen_details} for more details). 
The feature names indicate the sets $\mathcal{S}$, that we perform direct recourse actions on. 
The white dots indicate the median entanglement costs, and the box indicates the interquartile range.
In line with Proposition \ref{proposition:recourse_costs_ae}, the costs are pushed to $0$.
}
\label{fig:dis_error_prop1}
\end{figure*}

\subsection{Details on Experiments}\label{section:experimen_details}

\textbf{Real--world Data.} Our first data set is the \emph{Adult} data set taken from the UCI repository. This data set consists of approximately $48K$ samples with demographic (e.g., race, sex), education and employment (e.g., degree, occupation, hours-per-week), personal (e.g., marital status, relationship), financial (capital gain/loss) features where the label predicts whether an individual's income exceeds $50K\$$ per year ($y=1$). 
Our second data set, \emph{COMPAS} (Correctional Offender Management Profiling for Alternative Sanctions) consists of defendants' criminal history, jail and prison time, demographics and the goal is to predict recidivism risk for defendants from Broward County, Florida.
Our third data set is the \emph{Give Me Credit} data set from 2011 Kaggle competition.
It is is a credit scoring data set, consisting of 150,000 observations and 11 features. 
The classification task consists of deciding whether an instance will experience financial distress within the next two years (\textit{SeriousDlqin2yrs} is 0). 

\textbf{Prediction Models.} For all our experiments, we obtain counterfactual explanations for two classification models, for which we provide additional details in Appendix \ref{Appendix:architecture_params}:
We use is a binary logistic classifier that was trained without regularization, and an artificial neural network with a two-layer architecture that was trained with ReLU activation functions.

\textbf{Recourse Methods.}
For all data sets, recourses are generated in order to flip the prediction label from the unfavorable class ($y = 0$) to the favorable class ($y = 1$). 
We partition the data set into 80-20 train-test splits, and do the model training and testing on these splits. 
We use the following six methods as our baselines for comparison: 
\begin{itemize}
    \item[M:] \texttt{CLUE} \citep{antoran2020getting}: This model suggests feasible counterfactual explanations that are likely to occur under the data distribution. Using the VAE's decoder, \texttt{CLUE} uses an objective that guides the search of CEs towards instances that have low uncertainty measured in terms of the classifier's entropy. 
    \item[M:] \texttt{REVISE} \citep{joshi2019towards}: To find recourses that lie on the \emph{data manifold}, this method utilizes a trained autoencoder to transform the input space into a latent embedding space. \texttt{REVISE} then uses gradient descent in latent space to find recourses that lie on the data manifold.
    \item[M:] \texttt{CCHVAE} \citep{pawelczyk2019}: This is a method to find recourses that lie on the \emph{data manifold}. \texttt{CCHVAE} also uses a trained autoencoder to transform the input space into a latent embedding space. The latent representation is then randomly perturbed to find recourses.
    \item[G:] \texttt{FACE-K} \& \texttt{FACE-E} \citep{Poyiadzi2020}: This is classifier-agnostic method that finds recourses that lie on paths along dense regions. These methods construct \emph{neighbourhood graphs} to find paths through dense regions. The graph is either an $\epsilon$-graph (\texttt{FACE-E}) or a $k$-nearest neighbour graph (\texttt{FACE-K}).
\end{itemize}
Note that 
`M' abbreviates methods which generate recourses that lie on the data manifold, and `G' abbreviates methods, which use a graphical model to generate recourses that lead through dense paths. `D' refers to our method (i.e., \texttt{DEAR}), which takes input dependencies into account.
To allow for a fair comparison across the explanation models, which use autoencoders, we use similar base architectures for \texttt{DEAR}.
Appendix \ref{Appendix:architecture_params} provides implementation details for the recourse methods and of all used autoencoders. 
We compute evaluation measures by using the min-max normalized inputs used for training the classification and generative models. 
Below we describe the evaluation measures.


\textbf{Evaluation Measures.}
Since we are interested in generating small cost recourses, we define a notion of distance from the counterfactual explanation to the input point. 
As all methods under consideration minimize the $\ell_1$ norm, we use this measure and compare the $\ell_1$-costs across the methods. 
Further, we count the constraint violations (CV). 
We set the protected attributes `sex' and `race' to be immutable for the Adult and COMPAS data sets, and count how often each of the explanation models suggests changes to these protected features. 
GMC has no protected attribute.
We use the label yeighborhood (YNN) and measure how much data support recourses have from positively classified instances \citep{pawelczyk2021carla}.
Ideally, recourses should be close to correct positively classified individuals, which is a desideratum formulated by the authors of \citep{laugel2019dangers}.
Values of \text{YNN} close to 1 imply that the neighbourhoods around the recourses consists of points with the same predicted label, indicating that the neighborhoods around these points have already been reached by positively classified instances.
Note that some generated recoures do not alter the predicted label of the instance as anticipated. 
Therefore we keep track of the success rate, i.e., how often do the suggested counterfactuals yield successful recourse.
We so by counting the fraction of the respective methods' correctly determined counterfactuals.
Finally, we report the entanglement costs.
For a fixed set $\mathcal{S}$, for every instance at the end of training, we obtain $d$ Hessian matrices $\bH^{(j)} = \frac{\partial^2 g}{\partial \bv \partial x_S}$ for $1 \leq j \leq d$.
We then average the Hessian off-diagonal elements across all $j$ and plot their distribution across all training instances.
We can only do this for our recourse method \texttt{DEAR}.
 

\subsection{Experimental Results} \label{section:experiment_evaluation}
\textbf{Recourse Costs.}
The baseline comparisons regarding the cost of recourse are shown in Figure \ref{fig:cost_results}.
Relative to manifold-based recourse methods (i.e, \texttt{REVISE} and \texttt{CCHVAE}), \texttt{DEAR} usually performs more favourably ensuring up to 50 percent less costly median recourse costs. 
This is due to the fact that \texttt{DEAR} can use the most discriminative features in input space -- as opposed to latent space -- to search for recourses. 
Relative to the graph-based methods (i.e., \texttt{FACE}) our method performs significantly better.
Since \texttt{FACE} has to ensure connected paths their costs are usually highest.

\begin{figure}[tb]
\centering
\begin{subfigure}[b]{0.49\columnwidth}
\centering
\includegraphics[scale=0.32]{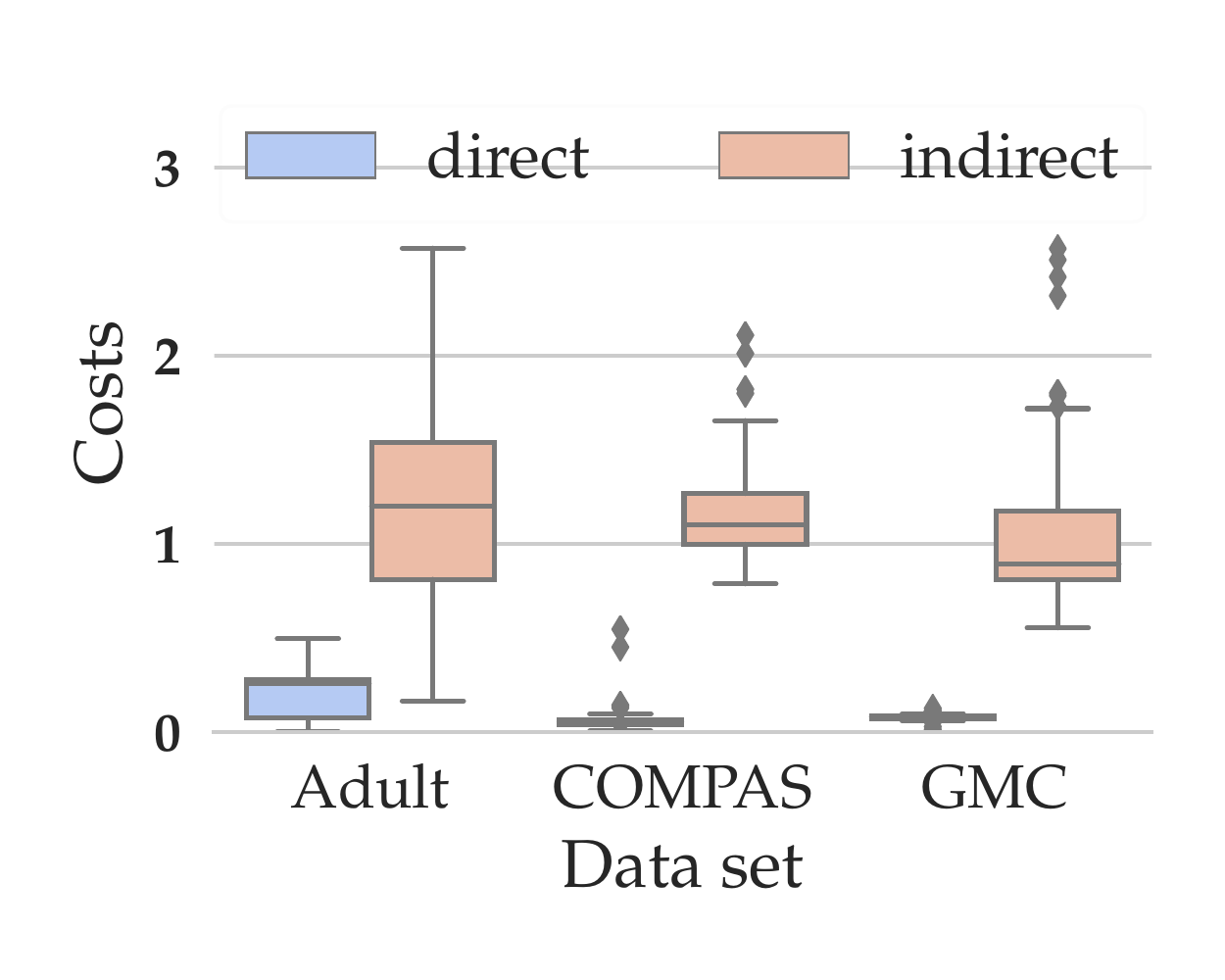}
\caption{Logistic Regression}
\label{fig:cost_split_linear}
\end{subfigure}
\hfill
\begin{subfigure}[b]{0.49\columnwidth}
\centering
\includegraphics[scale=0.32]{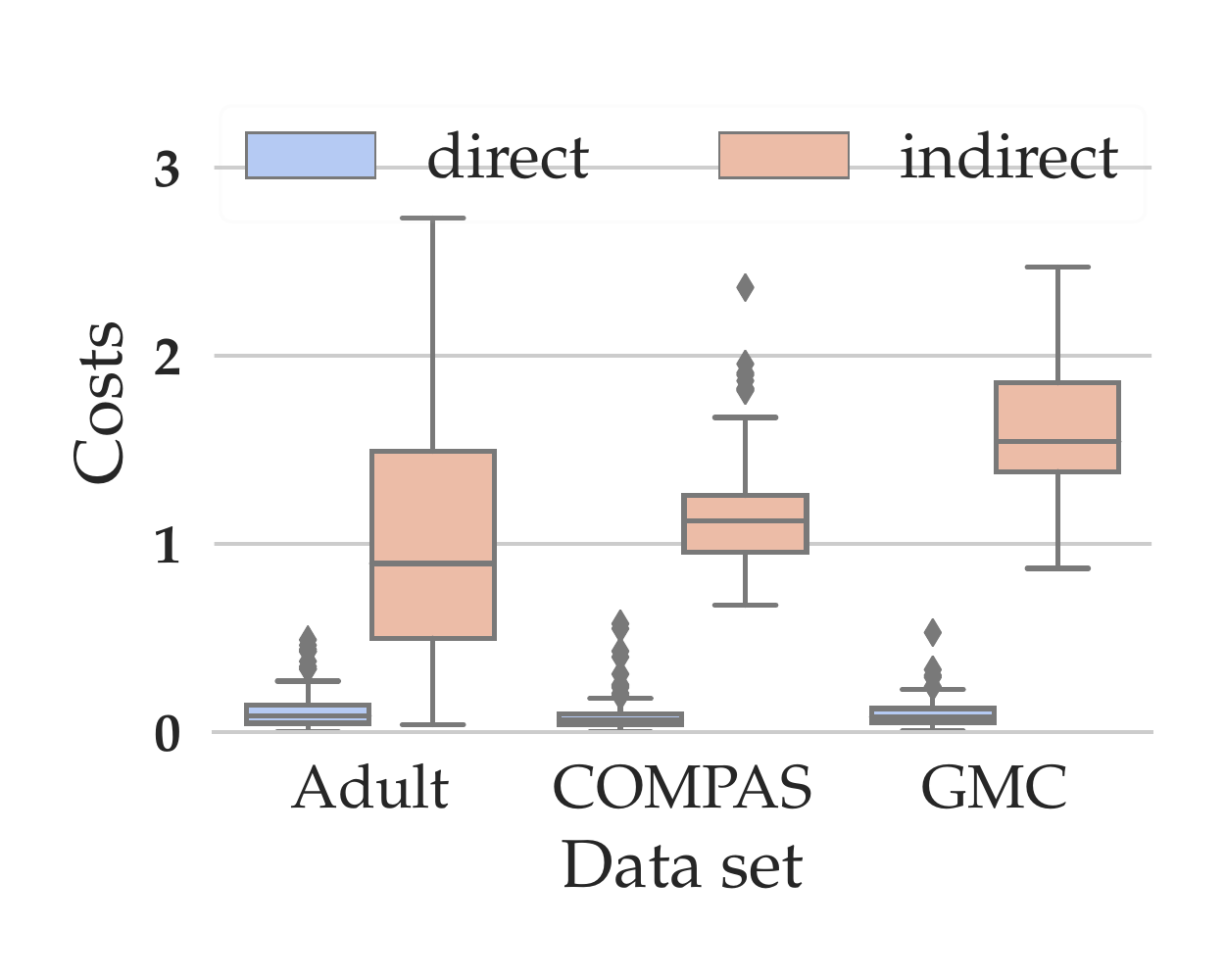}
\caption{Artificial Neural Network}
\label{fig:cost_split_ann}
\end{subfigure}
\caption{Cost splits as suggested by Proposition \ref{proposition:recourse_costs_ae} on both classifiers across all data sets.
The \textcolor{RoyalBlue}{direct costs} corresponds to the direct action \textcolor{RoyalBlue}{$\bd_{\mathcal{S}}$} and are measured as $\lVert \textcolor{RoyalBlue}{\bd_{\mathcal{S}}} \rVert_1$. 
The \textcolor{BrickRed}{indirect costs} are measured as $\lVert \bx_{S^c} -  \textcolor{BrickRed}{g_{\bx_{S^c}}}(\bv, \textcolor{RoyalBlue}{\bd_{\mathcal{S}}} + \bx_S) \rVert_1$.
}
\label{fig:cost_splits}
\end{figure}

\textbf{Reliability of Recourse.}
We measure the reliability of recourse using SR, CV and YNN presented in the previous Section. The results across all methods, data sets and classifiers are shown in Table \ref{tab:reliability}.
We see that \texttt{DEAR} has the highest SRs across all data sets and classifiers, among the highest YNN scores, and one of the lowest CV rates. 
Compared to the manifold-based recourse methods, \texttt{DEAR}'s SR is up to 45 percentage points higher. 
This is due to the fact that the lower dimensional data manifold can end before the decision boundary is reached and thus the manifold-based methods, which search for recourse in latent space, sometimes get stuck before they find a counterfactual instance (see \citep{downs2020interpretable} for a detailed analysis of this phenomenon). 
\citet[Appendix]{antoran2020getting} report a similar finding. 
A similar reason likely prevents \texttt{FACE} from reaching high SRs.
Our method does not suffer from this shortcoming since it primarily uses the most discriminative features in input space (Proposition \ref{proposition:dear}) to search for recourses, resulting in SRs of $1$.

\textbf{Qualitative Analysis.}
Finally, we analyze our recourse model qualitatively. 
We start by analyzing the \emph{entanglement costs}. 
As required by Proposition \ref{proposition:recourse_costs_ae}, we require these costs to be pushed to 0. 
We plot the distribution of the averaged off-diagonal terms in Figure \ref{fig:dis_error_prop1}. 
The results show that the entanglement cost is consistently pushed to 0 (most medians are at 0). 
These results indicate that our mechanism is very well aligned with Proposition \ref{proposition:recourse_costs_ae}'s requirement of disentangled $\bv$ and $\bx_S$.
Next, we analyze the cost splits. According to Proposition \ref{proposition:recourse_costs_ae}, we can split the costs of recourse into a direct and an indirect component. We show these cost splits in Figure \ref{fig:cost_splits} verifying that the elasticity of $g_{\bx_{S^c}}$ w.r.t.\ $\bx_{\mathcal{S}}$ is non-zero, i.e., we observe a strong presence of feature dependencies. In App.\ \ref{appendix:case_study} we provide a case study and further analyze the semantic meaning of these cost splits on GMC.

\section{Conclusion}
In this work, we considered the problem of generating algorithmic recourse in the presence of feature dependencies -- a problem previously only studied through the lens of causality.
We developed \texttt{DEAR} (DisEntangling Algorithmic Recourse), a novel recourse method that generates recourses by disentangling the latent representation of co-varying features from a subset of promising recourse features to capture some of the main practical desiderata: (i) recourses should adhere to feature dependencies without the reliance on hand-crafted causal graphical models and (ii) recourses should lie in dense regions of the feature space, while providing (iii) low recourse costs.
Quantitative as well as qualitative experiments on real-world data corroborate our theoretically motivated recourse model, highlighting our method's ability to provide reliable and low-cost recourse in the presence of feature dependencies. 

We see several avenues for future work. 
From an end-user perspective, comparing the practical usefulness across various different recourse methods running user-studies with human participants is an important direction for future work. 
Further, our framework showcases the importance of feature dependencies for reliable algorithmic recourse by highlighting which individual features contributed directly and indirectly to the recourse. 
While this is reminiscent of recourses output by causal methods the recourses output by our framework should not be mistaken for causal recourses. 
Therefore, from a theoretical perspective, it would be interesting to find (local) conditions for both the classifier and the generative model under which our recourse framework would generate recourses with a causal interpretation.

\bibliography{bib_main}
\bibliographystyle{plainnat}

\appendix

%
%





%

%

\onecolumn
\aistatstitle{Supplementary Materials: \\ Decomposing Counterfactual Explanations \\ for Consequential Decision Making}

\section{Theoretical Analysis} \label{appendix:proof}
\subsection{Proof of Proposition \ref{proposition:recourse_costs_ae}} \label{appendix:proof_disentanglement}
\emph{Proof}(Recourse Costs by \texttt{DEAR})
First, we note that $\bv$ is usually obtained via some kind of training procedure, and thus it could be a function of $\bx_S$. Next, we partition both $\bz=[ \bv ~~ \bx_S ]^\top$ and  $\bd_z=[\bd_v ~ \bd_{S} ]^\top$. Moreover, we partition $g(\bv, \bx_S)=[ g_{\bx_S}(\bv,\bx_S) ~~ g_{\bx_{S^c}} (\bv,\bx_S) ]^\top$ $= [ \bx_S ~  \bx_{S^c} ]^\top$. Then, the matrix of derivatives can be partitioned as follows:
 \begin{equation*}
     \bJ_z^{(\bx)} = \begin{bmatrix} \bJ_\bv^{(\bx_{S^c})} & \bJ_{x_S}^{(\bx_{S^c})} \\ \bJ_{\bv}^{(\bx_S)} & \bJ_{\bx_S}^{(\bx_S)} \end{bmatrix} :=  \begin{bmatrix} \bA & \bB \\ \bC & \bD \end{bmatrix}.
 \end{equation*}
 Since we are not interested in applied changes to $\bv$ we set $\bd_v{:=}\mathbf{0}$. By Lemma \ref{lemma:recourse_cost_general} we know $ \lVert \bd_x \rVert^2 {=}\bd^\top_z \big({\bJ_z^{(\bx)}}^\top \bJ_\bz^{(\bx)}\big) \bd_z$. A direct computation with $\bd_z {:=} \begin{bmatrix} \mathbf{0} & \bd_{S} \end{bmatrix}^\top$  yields:
 \begin{align*}
     \lVert \bd_x \rVert^2 &\approx \begin{bmatrix} \mathbf{0} & \bd_{S} \end{bmatrix} \begin{bmatrix} \bA & \bC \\ \bB & \bD \end{bmatrix} \begin{bmatrix} \bA & \bB \\ \bC & \bD \end{bmatrix} \begin{bmatrix} \mathbf{0} \\ \bd_{S} \end{bmatrix} \\
     & = \bd_{S}^\top \bB^\top \bB \bd_{S} + \bd_{S}^\top \bD^\top \bD \bd_{S} \\
     &=    \underbrace{\bd^\top_{S} \big({\bJ_{\bx_{S}}^{(\bx_{S^c})}}^\top \bJ_{\bx_S}^{(\bx_{S^c})}\big) \bd_{S}}_{\textcolor{BrickRed}{\text{Indirect Costs}}} + \underbrace{\bd^\top_{S} \big({\bJ_{\bx_S}^{(\bx_S)}}^\top \bJ_{\bx_S}^{(\bx_S)}\big) \bd_{S}}_{\textcolor{RoyalBlue}{\text{Direct Costs}}}.
\end{align*}
By the chain rule of multivariate calculus (recall that $\bv$ and $\bx_S$ need not be independent), note that we can write out the above terms as follows:
\begin{align*}
    {\bJ_{\bx_S}^{(\bx_{S^c})}} &= 
    \frac{\partial g_{\bx_{S^c}}(\bv, \bx_S)}{\partial \bv} \frac{\partial \bv}{\partial \bx_S} + \frac{\partial g_{\bx_{S^c}}(\bv, \bx_S)}{\partial \bx_S} \frac{\partial \bx_S}{\partial \bx_S} \\
    & = \underset{\textcolor{BrickRed}{\text{Entanglement costs}}}{\underbrace{\frac{\partial g_{\bx_{S^c}}(\bv, \bx_S)}{\partial \bv} \frac{\partial \bv}{\partial \bx_S}}} + \underset{\textcolor{BrickRed}{\text{Elasticity of $g{(\bx_{S^c})}$  w.r.t to $\bx_S$} }}{\underbrace{\frac{\partial g_{\bx_{S^c}}(\bv, \bx_S)}{\partial\bx_S}}} \\
    {\bJ_{\bx_S}^{(\bx_S)}} &= 
    \frac{\partial g_{\bx_{S}}(\bv, \bx_S)}{\partial\bv} \frac{\partial\bv}{\partial\bx_S} + \frac{\partial g_{\bx_{S}}(\bv, \bx_S)}{\partial\bx_S} \frac{\partial \bx_S}{\partial\bx_S} \\
    &= \underset{\textcolor{RoyalBlue}{\text{Entanglement costs}}}{\underbrace{\frac{\partial g_{\bx_{S}}(\bv, \bx_S)}{\partial\bv} \frac{\partial\bv}{\partial\bx_S}}} + \underset{\textcolor{RoyalBlue}{\text{Identity Mapping}}}{\underbrace{\frac{\partial g_{\bx_{S}}(\bv, \bx_S)}{\partial\bx_S}}}.
   \tag*{\qed}
\end{align*}
 Let us consider what this implies intuitively. 
 For the direct costs, notice that $g$ would achieve the best reconstruction of $\bx_S$ by using the identify mapping. Recall, in Section \ref{section:marie_algorithm}, we suggested to use a \texttt{ResNet} component within the decoder to enforce this identity mapping during training of our autoencoder model. Hence, under perfect disentanglement the disentanglement costs are 0, and the $\bJ_{\bx_S}^{(\bx_S)}{=} \mathbf{1}$: thus, the direct cost would ideally be given by $\bd_{S}^{\top} \bd_{S}$. This is the squared $\ell_2$ norm of $\bd_{S}$. 
 
 The indirect costs, on the other hand, depend on the sensitivity of $\bx_{S^c}$ with respect to $\bx_S$, that is, $\bJ_{\bx_{S}}^{(\bx_{S^c})}$. Again, we consider the case of perfect disentanglement first: Suppose $\bx_S$ was a variable that was unrelated to the remaining variables $\bx_{S^c}$, while still being predictive of the outcome: Then $\bJ_{x_{S}}^{(x_{S^c})} = \mathbf{0}$, and a change $\bd_{S}$ would only have a direct impact on the outcome, and thus the indirect costs would disappear. In this case, the recourse cost for independence--based and dependence--based methods coincide. On the other extreme, suppose $\bx_{S}$ was almost a copy of $\bx_{S^c}$, then $\bJ_{\bx_{S}}^{(\bx_{S^c})} \approx \mathbf{1}$, and changing $\bx_S$ clearly impacts the remaining variables $\bx_{S^c}$. In this case, an independence--based method would not reliably capture the recourse costs. 

\subsection{Proof of Proposition \ref{proposition:dear}} \label{appendix:proof_closed_form_dear}

\begin{proof}[Proof of Proposition \ref{proposition:dear}]
Recall the problem in \eqref{equation:problem_dependent_partition}:
\begin{equation}
\bd_{\mathcal{S}}^* = \argmin_{\bd_{\mathcal{S}}} \mathcal{L} = \argmin_{\bd_{\mathcal{S}}} \lambda \lVert \bd_{\mathcal{S}} \rVert^2  + \lVert s - f\big(g(\bv, \bx_{\mathcal{S}} + \bd_{\mathcal{S}})\big)  \rVert^2.
\label{equation:objective_proof_dependent}
\end{equation}
We use the following first-order approximation to $f\big(g(\bv, \bx_{\mathcal{S}} + \bd_{\mathcal{S}})\big) \approx f(\bx) + \nabla f(\bx)^\top {\bY_{\bx_{\mathcal{S}}}^{(\bx)}} \bd_{\mathcal{S}}$, where we have substituted $g(\bv, \bx_{\mathcal{S}}) = \bx$ and used that $\bv \upmodels \bx_{\mathcal{S}}$ by design of the generative model. Then, we can derive a surrogate loss to the loss from \eqref{equation:objective_proof_dependent}:
 \begin{align}
\mathcal{L} \approx \tilde{\mathcal{L}} = \lambda \cdot \bd_{\mathcal{S}}^\top \bd_{\mathcal{S}} +  \big(m - \nabla f(\bx)^\top {\bY_{\bx_{\mathcal{S}}}^{(\bx)}} \bd_{\mathcal{S}} \big)^\top \big( m - \nabla f(\bx)^\top {\bY_{\bx_{\mathcal{S}}}^{(\bx)}} \bd_{\mathcal{S}} \big),
\label{equation:surrogate_loss}
\end{align}
where $m = s - f(\bx)$.
The second term on the right in \eqref{equation:surrogate_loss} can be written as:
\begin{align*}
m^2 - 2 m \nabla f(\bx)^\top {\bY_{\bx_{\mathcal{S}}}^{(\bx)}} \bd_{\mathcal{S}} + \bd_{\mathcal{S}}^\top  {\bY_{\bx_{\mathcal{S}}}^{(\bx)}}^\top \nabla f(\bx) \nabla f(\bx)^\top  {\bY_{\bx_{\mathcal{S}}}^{(\bx)}} \bd_{\mathcal{S}}.
\end{align*}
By solving $\argmin_{\bd_{\mathcal{S}}} \tilde{\mathcal{L}}$ we find the optimal change required in the features $\bx_{\mathcal{S}}$ as follows:
\begin{align}
    \tilde{\bd}_{\mathcal{S}}^* = \bm{M}^{-1}  \bm{\eta},
\end{align}
where
\begin{align}
\bm{M} & = \lambda \cdot \bI + {\bY_{\bx_{\mathcal{S}}}^{(\bx)}}^\top \nabla f(\bx) \nabla f(\bx)^\top  {\bY_{\bx_{\mathcal{S}}}^{(\bx)}}  &
\bm{\eta} & = m \cdot \nabla f(\bx)^\top {\bY_{\bx_{\mathcal{S}}}^{(\bx)}}.
\label{eq:closed_form}
\end{align}
Next, we define $\bw = {\bY_{\bx_{\mathcal{S}}}^{(\bx)}}^\top \nabla f(\bx)$. Note that $\bw \bw^\top$ is a rank-1 matrix. Thus, by the well-known Sherman-Morrison-Woodbury formula, $\bM$ can be inverted as follows:
\begin{align}
    \bM^{-1} = \frac{1}{\lambda} \bigg(\bI - \frac{\bw \bw^\top}{\lambda + \lVert \bw \rVert_2^2} \bigg).
\label{eq:inverse}
\end{align}
As a consequence, after substituting \eqref{eq:inverse} into \eqref{eq:closed_form} we obtain that:
\begin{align}
    \tilde{\bd}_{\mathcal{S}}^* = \frac{m}{\lambda + \lVert \bw \rVert_2^2} \bw.
\end{align}
Further, note that $\bdelta = g(\bz + \bd_{\mathcal{S}}) - g(\bz) \approx {\bY_{\bx_{\mathcal{S}}}^{(\bx)}} \bd_{\mathcal{S}}$, where we have used that $\bv \upmodels \bx_{\mathcal{S}}$.
Therefore, we obtain a first-order approximation to the optimal recourse in input space:
\begin{align}
\bdelta^*_x \approx {\bY_{\bx_{\mathcal{S}}}^{(\bx)}} \tilde{\bd}_{\mathcal{S}}^* = \frac{m}{\lambda + \lVert \bw \rVert_2^2} \cdot {\bY_{\bx_{\mathcal{S}}}^{(\bx)}} \bw,
\label{eq:closed_form_explicit}
\end{align}
as claimed.
\end{proof}

\subsection{Proof of Lemma 1}
\begin{lemma}[Recourse costs in terms latent space quantities]
Given a latent representation $\bz$ of a sample $\bx=g(\bz)$ and a generated counterfactual $\xC=g(\zC)$ with $\zC = \bz + \bd_z$, the cost of recourse $\lVert \bx - \xC \rVert^2$ can be expressed in terms of latent space quantities:
\begin{align*}
 \lVert \bdelta_x \rVert^2
 &\approx   \bd^\top_z \big({\bJ_z^{(\bx)}}^\top \bJ_z^{(x)}\big) \bd_z ,
\end{align*}
where the matrix of derivatives with respect to output $\bx=g(\bz)$ is given by $\left.\bJ_z^{(\bx)}{:=} \frac{\partial g(\bz)}{\partial \bz }\right\vert_{\bz=\bz}$.
\label{lemma:recourse_cost_general}
\end{lemma}
\begin{proof}
We use the cost of recourse, and a first-order Taylor series approximation for $g(\bz+ \bd_z)$ at $\bz$ to arrive at:
\begin{align*}
\lVert \bdelta_x \rVert^2 &= \lVert g(\bz) -g(\bz + \bd_z) \rVert^2  \\
&\approx \lVert g(\bz) - (g(\bz)+{\bJ_\bz^{(\bx)}}\bd_z) \rVert^2  \\
 &= \bd^\top_z \big({\bJ_z^{(\bx)}}^\top \bJ_z^{(\bx)}\big) \bd_z,
\end{align*}
where $\left.\bJ_\bz^{(\bx)} := \frac{\partial g(\bz) }{\partial \bz }\right\vert_{\bz=\bz}$.
\end{proof}
On an intuitive level, Lemma \ref{lemma:recourse_cost_general} measures how the cost -- measured in input space quantities -- depends on perturbations of each component of the generative latent space $\bz$.

\section{Case Study: ``Credit Risk''}\label{appendix:case_study}
\begin{figure}[tb]
\centering
\begin{subfigure}[b]{0.49\columnwidth}
\centering
\includegraphics[width=\columnwidth]{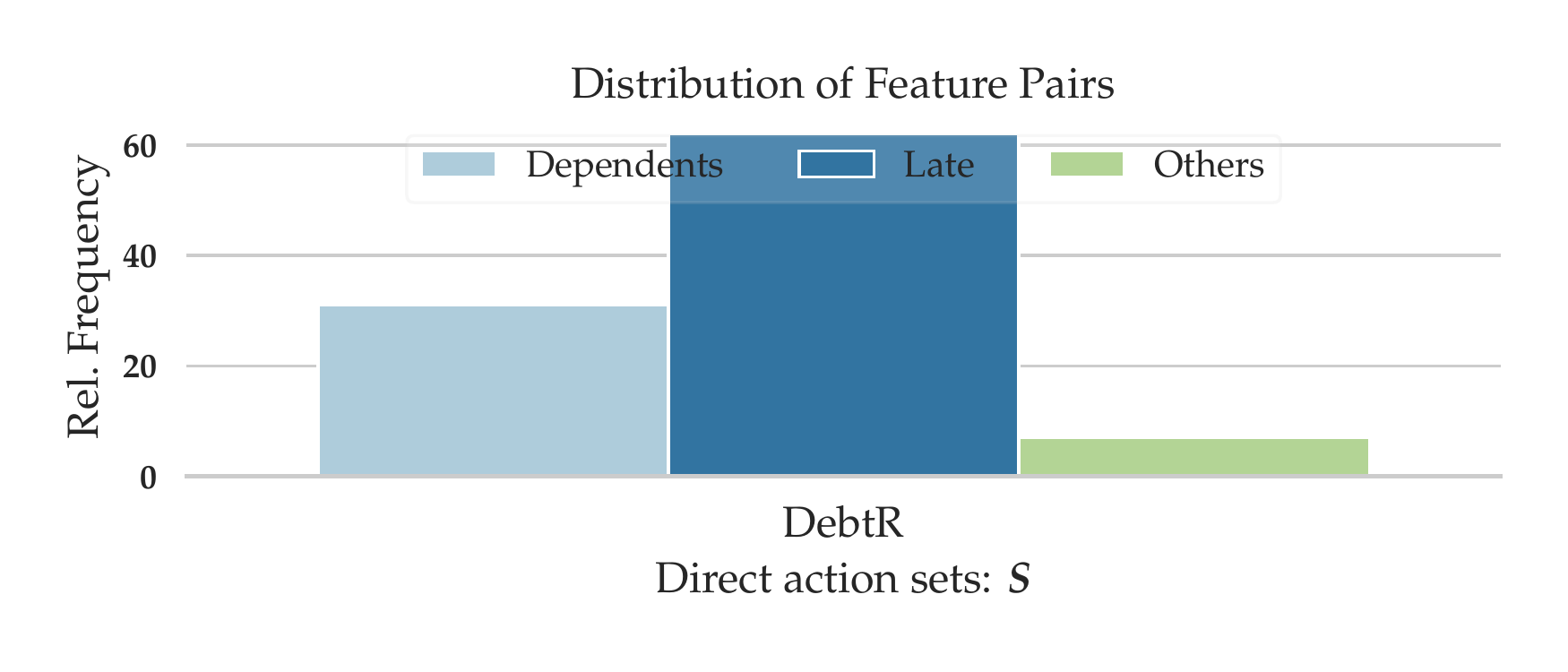}
\caption{Logistic Regression}
\label{fig:cost_split_gmc_lr}
\end{subfigure}
\hfill
\begin{subfigure}[b]{0.49\columnwidth}
\centering
\includegraphics[width=\columnwidth]{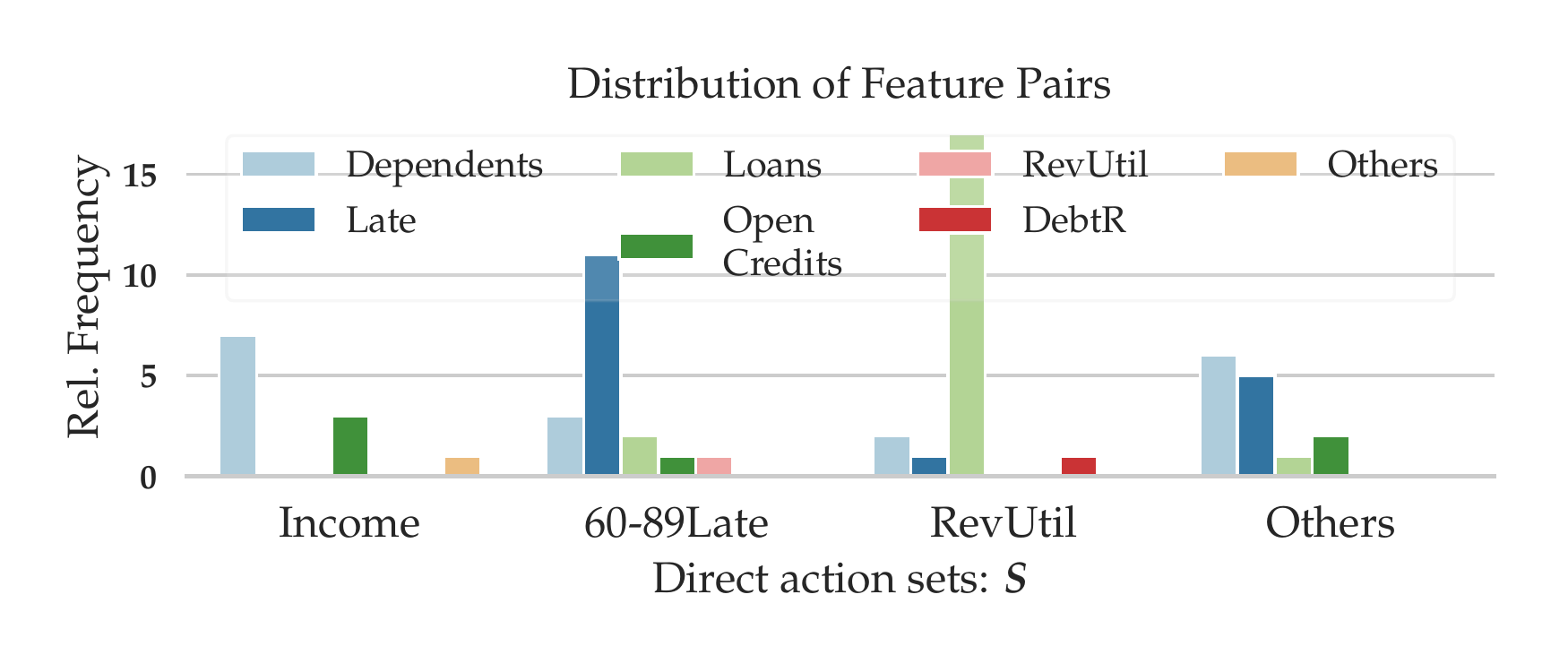}
\caption{Artificial Neural Network}
\label{fig:cost_split_gmc_ann}
\end{subfigure}
\caption{We show the rel.\ frequency of important feature pairs that need to be changed together.
Associated with each direct minimum cost action on the x-axis (i.e., $\bd_{\mathcal{S}}$), we plot the second most important feature (y-axis) that should change together with the direct action feature from $\mathcal{S}$.
For example, in the bottom panel, for 10 percent of all instances, decreasing '60-89 days late' goes hand in hand with either a decrease in '30-60 days late' or a decrease in 'more than 90 days late'.}
\label{fig:cost_splits_gmc}
\end{figure}

\begin{figure}
\begin{subfigure}[b]{\textwidth}
\centering
\includegraphics[width=\textwidth]{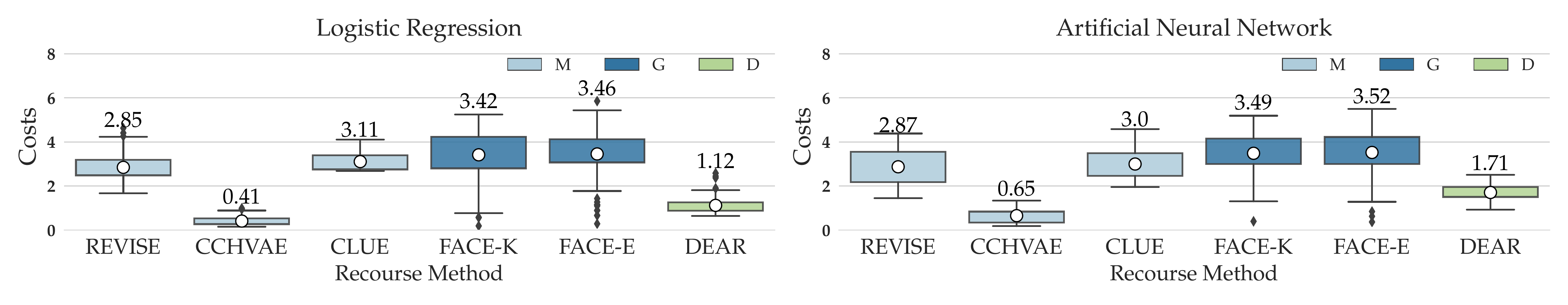}
\end{subfigure}
\caption{Give Me Credit: Plot from main text.}
\label{fig:gmc_cost_comparison}
\end{figure}

As a practical example, we showcase additional insights that our recourse model can provide.
Here we analyze the cost splits from a semantic point of view.
In Figure \ref{fig:cost_splits_gmc}, we show the distribution of important feature pairs that need to change together to lead to loan approvals for individuals from the \emph{Give Me Credit} data set: 
the x-axis shows the direct actions resulting in the lowest costs, and the y-axis shows the relative frequency of the most important indirect actions.
The following noteworthy patterns emerge: 
(i) the non-linear classifier has picked up more non-linear relations since the feature, on which the minimum cost direct actions are suggested, vary more heavily across instances for the ANN model (bottom panel) relative to the LR model (top panel); 
For example, a decrease in `revolving utilization' is often followed by a decrease in the number of `loans', which is semantically meaningful suggesting ways to reduce the `revolving utilization'.
Finally, we emphasize that our method showcases the importance of feature dependencies for reliable algorithmic recourse by highlighting how it arrived at the recourse.
The recourses output by our framework should not be mistaken for causal recourses.

\begin{table}[tb]
\centering
\begin{tabular}{lrccc}
\toprule
& & \multicolumn{3}{c}{Recourse Method} \\
\cmidrule(lr){3-5}
\multicolumn{2}{c}{Factual} & \texttt{SCFE} (IMF) & \texttt{REVISE} (M) & \texttt{DEAR} (D) \\ 
\cmidrule(lr){1-2} \cmidrule(lr){3-3}
\cmidrule(lr){4-4}
\cmidrule(lr){5-5}
\texttt{Age} & 45 & \textbf{58} & 45 & 45  \\
\texttt{Educ} & 9 & \textbf{7} & \textbf{11} & 9 \\
\texttt{C-gain} & 0 & 0 & \textbf{2110} & \textcolor{BrickRed}{\textbf{307}} \\
\texttt{Hours} & 0 & 0 & \textbf{80} & \textcolor{BrickRed}{\textbf{24}} \\
\texttt{C-loss} & 60 & \textbf{20} & \textbf{45} & \textcolor{RoyalBlue}{\textbf{41}} \\
\texttt{W-class} & Private: No & No & No &No \\
\texttt{M-status} & Married: No & No & No & No \\
\texttt{Occu} & Specialized: No & No & \textbf{Yes} & \textcolor{BrickRed}{\textbf{Yes}} \\
\texttt{Race} & White: No & No& No& No\\
\texttt{Sex} & Male: No  & No & No &No \\
\texttt{Native} & US: Yes & Yes & Yes & Yes \\
\midrule
ANN & $>50K$: No & \textbf{Yes} & \textbf{Yes} & \textbf{Yes} \\
\bottomrule
\end{tabular}%
\caption{Comparing recourses qualitatively across explanation methods for a factual instance. The bottom row shows the ANN classifier's predicted class outcomes. For \texttt{DEAR}, the recourse intervention was done to \textcolor{RoyalBlue}{\texttt{C-loss}}. The remaining features changed as a result of this. For \texttt{CCHVAE}, the results were similar to those by \texttt{REVISE}.}
\label{tab:typical_example}
\end{table}

\section{Implementation Details} \label{Appendix:architecture_params}

\subsection{Handling Constraints}
\textbf{Encoding Monotonicity Constraints.}
In the presence of strong prior knowledge on how certain features are allowed to change (e.g., `years of schooling' (yos) or `age' can only go up) one can add Hinge-losses \citep{mahajan2019preserving} to encourage monotonicity constraints.
Let $x_{yos}$ correspond to the schooling feature. Then we can add $- \min(0, \check{x}_{yos} - x_{yos})$ to the loss function in \eqref{equation:surrogate_loss_dear} to ensure that the counterfactual $\check{x}_{yos}$ should increase, where $\check{x}_{yos}$ is the corresponding entry from $g(\bv, \bx_S+\bd_{\mathcal{S}})$. 

\textbf{Handling Categorical Variables.}
Using \texttt{DEAR}, one can easily handle (high-cardinality) categorical features. We can turn all categorical features into numeric features by standard one-hot encoding. For each categorical feature, we can then use a softmax-layer after the final output layer of the decoder. 
For the purpose of the one-hot-encoded reconstruction, we apply the argmax.



\subsection{Recourse Methods}
For all data sets, the features are binary-encoded and the data is scaled to lie between $0$ and $1$. 
We partition the data sets into train-test splits. The training set is used to train the classification models for which recourses are generated.
Recourses are generated for all samples in the test split for the fixed classification model. 
In particular, we use the following algorithms to generate recourses. Specifically,
\begin{itemize}


\item \underline{{\texttt{C-CHVAE}}} An autoencoder is additionally trained to model the data-manifold. 
The explanation model uses a counterfactual search algorithm in the latent space of the AE. 
Particularly, a latent sample within an $\ell_1$-norm ball with search radius $r_l$ is used until recourse is successfully obtained. 
The search radius of the norm ball is increased until recourse is found. 
The architecture of the generative model are provided in Appendix~\ref{app:vae_architecture}.

\item \underline{{\texttt{REVISE}}} As with the recourse model of \citet{pawelczyk2019}, an autoencoder is additionally trained to model the data-manifold. The explanation model uses a gradient-based search algorithm in the latent space of the AE.
For a fixed weight on the distance component, we allow up to 500 gradient steps until recourse is successfully obtained.
Moreover, we iteratively search for the weight leading up to minimum cost recourse. 
The architectures of the generative model are provided in Appendix~\ref{app:vae_architecture}.


\item{\underline{\texttt{FACE}}}
\citet{Poyiadzi2020} provide \texttt{FACE}, which uses a shortest path algorithm (for graphs) to find counterfactual explanations from  high--density regions. Those explanations are actual data points from either the training or test set. Immutability constraints are enforced by removing incorrect neighbors from the graph. We implemented two variants of this model: one uses an epsilon--graph (\texttt{FACE-EPS}), and a second one uses a knn--graph (\texttt{FACE-KNN}). To determine the strongest hyperparameters for the graph size we conducted a \emph{grid search}. We found that values of $k_{\text{FACE}}=50$ gave rise to the best balance of success rate and costs. For the epsilon graph, a radius of 0.25 yields the strongest results to balance between high $ynn$ and low cost.

\item{\underline{\texttt{CLUE}}} \citet{antoran2020getting} propose \texttt{CLUE}, a generative recourse model that takes a classifier's uncertainty into account. This model suggests feasible counterfactual explanations that are likely to occur under the data distribution. The authors use a  variational autoencoder (VAE) to estimate the generative model. Using the VAE's decoder, \texttt{CLUE} uses an objective that guides the search of CEs towards instances that have low uncertainty measured in terms of the classifier's entropy. We use the default hyperparameters, which are set as a function of the data set dimension $d$. Performing hyperparameter search did not yield results that were improving distances while keeping the same success rate.

\end{itemize}

We describe architecture and training details in the following.

\subsection{Supervised Classification Models}

All models are implemented in PyTorch and use a $80-20$ train-test split for model training and evaluation. We evaluate model quality based on the model accuracy. All models are trained with the same architectures across the data sets:

\begin{table}[ht]
\centering
\begin{tabular}{@{}lll@{}}
\toprule
                         & Neural Network                       & Logistic Regression        \\ \midrule
Units                    & [Input dim, 18, 9, 3, 1] & [Input dim, 1] \\
Type                     & Fully connected                      & Fully connected            \\
Intermediate activations & ReLU                                & N/A                        \\
Last layer activations   & Sigmoid                              & Sigmoid                    \\ \bottomrule
\end{tabular}
\caption{Classification Model Details}
\label{tab:model_detils}
\end{table}

\begin{table}[ht]
\centering
\begin{tabular}{@{}l|llll@{}}
\toprule
              &                                                               & Adult & COMPAS & Give Me Credit \\ \midrule
Batch-size    & ANN                                                            & 512   & 32     & 64            \\ \cmidrule(l){2-5} 
              & \begin{tabular}[c]{@{}l@{}}Logistic\\ Regression\end{tabular} & 512   & 32     & 64            \\ \midrule
Epochs        & ANN                                                            & 50    & 40     & 30            \\ \cmidrule(l){2-5} 
              & \begin{tabular}[c]{@{}l@{}}Logistic\\ Regression\end{tabular} & 50    & 40     & 30            \\ \midrule
Learning rate & ANN                                                            & 0.002 & 0.002  & 0.001         \\ \cmidrule(l){2-5} 
              & \begin{tabular}[c]{@{}l@{}}Logistic\\ Regression\end{tabular} & 0.002 & 0.002  & 0.001         \\ \bottomrule
\end{tabular}
\caption{Training details}
\label{tab:train_dets}
\end{table}

{\begin{table}[ht]
\centering
\begin{tabular}{@{}lccc@{}}
\toprule
 & Adult & COMPAS & Give Me Credit \\ \midrule
Logistic Regression & 0.83  & 0.84  & 0.92  \\
Neural Network & 0.84  & 0.85  & 0.93  \\ \bottomrule
\end{tabular}
\caption{Performance of classification models used for generating algorithmic recourse.}
\label{tab:classifier_acc}
\end{table}
}

\subsection{Generative Model Architectures used for \texttt{DEAR}, \texttt{CCHVAE} and \texttt{REVISE}}\label{app:vae_architecture}
For all experiments, we use the following architectures.

\begin{table}[ht]
\centering
\begin{tabular}{@{}lccc@{}}
\toprule
& Adult                      & COMPAS  & Give Me Credit      \\ \midrule
Encoder layers                 & [input dim, 16, 32, 10] & [input dim, 8, 10, 5] & [input dim, 8, 10, 5]  \\
Decoder layers                  & [10, 16, 32, input dim] & [5, 10, 8, input dim] & [5, 10, 8, input dim] \\
Type                     & Fully connected                      & Fully connected & Fully connected \\
Loss function   & MSE                             & MSE  & MSE                 \\ \bottomrule
\end{tabular}
\caption{Autoencoder details}
\label{tab:autoencoder_details}
\end{table}

Additionally, for \texttt{DEAR} all generative models use the Hessian Penalty \citep{peebles2020hessian} and a residual block, which we both described in more detail in Section \ref{section:marie_algorithm} of the main text.

\end{document}